\documentclass{article} 
\usepackage{iclr2025_conference,times}

\usepackage{hyphenat}

\usepackage{ifthen}
\usepackage{amsmath,amsthm,amsfonts,amssymb,amscd,mathtools,bm,dsfont,empheq}

\DeclarePairedDelimiter\bracks{[}{]}
\DeclarePairedDelimiter\braces{\{}{\}}
\DeclarePairedDelimiterX{\bracesg}[2]{\{}{\}}{#1\:\delimsize|\:#2}

\DeclarePairedDelimiter\parens{\lparen}{\rparen}
\DeclarePairedDelimiterX{\parensg}[2]{\lparen}{\rparen}{#1\:\delimsize|\:#2}

\DeclarePairedDelimiter{\abs}{\lvert}{\rvert}
\DeclarePairedDelimiter{\norm}{\lVert}{\rVert}
\DeclarePairedDelimiterX{\divx}[2]{\lparen}{\rparen}{#1\:\delimsize\|\:#2}
\newcommand{\KL}{\operatorname{KL}\divx}

\ExplSyntaxOn
\cs_new_eq:NN \TextUppercase \text_uppercase:n
\cs_new_eq:NN \TextLowercase \text_lowercase:n
\ExplSyntaxOff

\let\vec\undefined
\newcommand{\blank}{\cdot}
\newcommand{\set}[1]{
\ifcat\noexpand#1\relax{#1}
\else\mathcal{\TextUppercase{#1}}\fi
}
\newcommand{\vec}[1]{\bm{\TextLowercase{#1}}}
\newcommand{\mat}[1]{\bm{\TextUppercase{#1}}}
\newcommand{\fun}[1]{\mathrm{#1}}

\newcommand{\call}[1]{\parens*{#1}}
\newcommand{\callg}[2]{\parensg*{#1}{#2}}
\newcommand{\bcall}[1]{\bracks*{#1}}

\newcommand{\Reals}[1][]{
  \ifthenelse{ \equal{#1}{} }
  {\ensuremath{\mathbb{R}}}
  {\ensuremath{\mathbb{R}^{#1}}}
}

\newtheorem{theorem}{Theorem}
\newtheorem{lemma}[theorem]{Lemma}     
\newtheorem{definition}{Definition} 

\newtheorem{proposition}[theorem]{Proposition}

\newtheorem{assumption}{Condition}

\newcommand{\E}[2][]{
  \ifthenelse{\equal{#1}{}}
  {\ensuremath{\operatorname{\mathbb{E}}\bracks*{#2}}}
  {\ensuremath{\operatorname{\mathbb{E}}_{#1}\bracks*{#2}}}
}

\definecolor{emph}{HTML}{ff5e5e}
\definecolor{true}{HTML}{006275}
\definecolor{proxy}{HTML}{9B2226}

\newcommand{\pidagger}{{ \pi^\dagger}}
\newcommand{\phidagger}{{\phi^\dagger}}

\newcommand{\pidaggertheta}{{ \pi^\dagger_\theta}}

\newcommand{\pibardagger}{{\bar{\pi}^\dagger}}
\newcommand{\pibardaggertheta}{{\bar{\pi}^\dagger_\theta}}
\newcommand{\ldagger}{{{\fun L}^\dagger}}

\newcommand{\pibar}{\bar{\pi}}

\newcommand{\pitilde}{{ \tilde\pi}}
\newcommand{\phitilde}{{\tilde{\phi}}}
\newcommand{\tautilde}{{\tilde{\tau}}}
\newcommand{\pitildetheta}{{ \tilde{\pi}_\theta}}
\newcommand{\phitildetheta}{{\tilde{\phi}_\theta}}
\newcommand{\tautildetheta}{{\tilde{\tau}_\theta}}

\newcommand{\ltilda}{{\tilde{\fun L}}}

\newcommand{\true}[1]{#1^\dagger}
\newcommand{\ctrue}[1]{{\color{true} #1}}

\newcommand{\cproxy}[1]{{\color{proxy} #1}}
\let\P\undefined
\newcommand{\P}{\mathcal{P}}

\usepackage{hyperref}
\usepackage{url}
\usepackage{todonotes}
\usepackage{tcolorbox}
\usepackage{import}
\usepackage{enumitem}
\usepackage{mathtools}

\title{When Can Proxies Improve the Sample Complexity of Preference Learning?}

\newcommand{\nextAuthor}{\hspace{3em}}

\author{
\begin{tabular}{c}
Yuchen Zhu$^1$%
\thanks{Correspondence to: \texttt{\scriptsize yuchen.zhu.18@ucl.ac.uk}}\nextAuthor
Daniel Augusto de Souza$^1$\nextAuthor
Zhengyan Shi$^1$\nextAuthor
Mengyue Yang$^2$ \\[1mm]
Pasquale Minervini$^3$\nextAuthor
Alexander D'Amour$^4$\nextAuthor
Matt J. Kusner$^1$ \\[1mm]
\end{tabular}\\
$^1$University College London,
$^2$University of Bristol,
$^3$University of Edinburgh,
$^4$Google Deepmind
}

\iclrfinalcopy 
\begin{document}

\maketitle

\begin{abstract}
We address the problem of \emph{reward hacking}, where maximising a proxy reward does not necessarily increase the true reward. This is a key concern for Large Language Models (LLMs), as they are often fine-tuned on human preferences that may not accurately reflect a true objective. Existing work uses various tricks such as regularisation, tweaks to the reward model, and reward hacking detectors, to limit the influence that such proxy preferences have on a model. Luckily, in many contexts such as medicine, education, and law, a sparse amount of expert data is often available. In these cases, it is often unclear whether the addition of proxy data can improve policy learning. We outline a set of sufficient conditions on proxy feedback that, if satisfied, indicate that proxy data can provably improve the sample complexity of learning the ground truth policy. These conditions can inform the data collection process for specific tasks. The result implies a parameterisation for LLMs that achieves this improved sample complexity. We detail how one can adapt existing architectures to yield this improved sample complexity.
\end{abstract}

\section{Introduction}

Large Language Models (LLMs) and other large generative models have revolutionised modern machine learning with their surprising capabilities, surpassing human-level performance in law, medicine, and other examinations \citep{achiam2023gpt,amin2023accuracy}. A large part of their success is their ability to incorporate human preferences to learn complex objectives such as trustworthiness \citep{yu2024rlhf}, sentiment preferences \citep{chakraborty2024maxmin}, and value alignment \citep{ji2023ai}. 

\begin{figure}[t!]
    \centering
    \includegraphics[width=1.\linewidth]{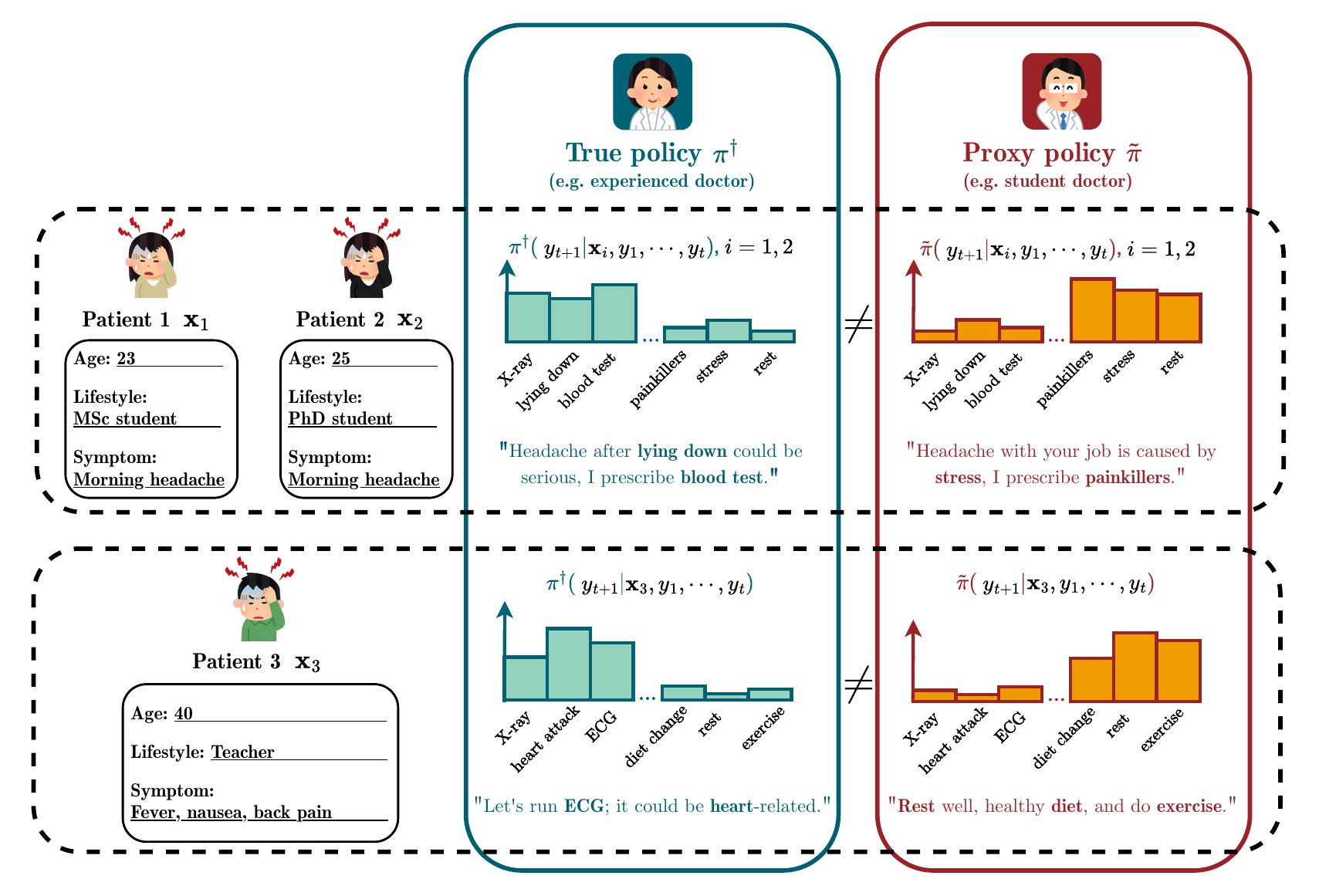}
    \caption{Medical question answering. (Illustrative purpose only. Not medical advice.) Patients 1 and 2 are put in the same group by both doctors as their key characteristics - age, lifestyle and symptom are all similar. Only the expert doctor correctly identifies that morning headache deserves a further check than headache at any other times in the day, since it could be caused by nerve in the brain pressured by a tumour. The student doctor naively attributes this to stress. Patient 3 has characteristics sufficiently different from Patient 1 and 2, so is put in a different group, and again the recommendations made by the two doctors are different.
    }
    \label{fig:motivating-example}
\end{figure}

In many cases, this preference data is a \emph{proxy} for the ground truth. For example, humans raters tend to prefer longer answers to a question, even if the answer is less informative \cite{zhou2024nature}. In this case, `response length' is a \emph{proxy} for the true helpfulness of an answer. If an LLM is trained on this proxy data alone it leads to a `length-bias' \citep{shen2023loose,singhal2023long}, as LLMs fine-tuned with this preference data generate longer and better formatted responses to appear more helpful \citep{chen2024odin}. This is an example of the well-known phenomenon of \emph{reward hacking}\footnote{This is also sometimes referred to as \emph{reward over-optimisation}.}: a model optimised to perform well with respect to a proxy reward function, performs poorly with respect to a ground truth reward function \citep{casper2023open}. Reward hacking is a fundamental problem in learning that has been observed in optimised circuits listening in on the oscillators of other computers when instead tasked to build their own \citep{bird2002evolved}, universities rejecting the most qualified applicants to boost their ratings \citep{golden2001glass}, and many other cases in game playing \citep{clark2016}, autonomous driving \citep{knox2023reward}, and text summarisation \citep{paulus2017deep}.

To address reward hacking in LLMs, prior work largely designs tweaks to the model, data, and optimization procedure. This includes regularisation towards an initial policy \citep{schulman2017proximal,rafailov2023direct,huang2024correctingmythos}, changing properties of the reward model \citep{gao2023scaling,coste2024reward}, using soft labels \citep{zhu2024iterative}, adjusting optimization hyperparameters \citep{singhal2023long}, reward hacking detection mechanisms \citep{pan2022effects,miao2024mitigating}, and introducing additional tools specialised to counteract length bias \citep{chen2024odin}. The reasoning behind this comes from the makeup of proxy data. We can think of proxy data as having two parts: (i) a \emph{true} part that brings a policy closer to the ground truth policy during learning and (ii) a \emph{false} part that moves it farther away. Prior work limits learning to reduce the impact that the false part has on the final model.

Without any further information on proxy preferences or the ground truth, we are restricted to methods such as these, i.e., methods that are blind to the true and false parts of proxy data, to reduce the impact of reward hacking \citep{pan2022effects}. Luckily, in many settings, we also have access to sparse observations of high-quality preferences \citep{daniels2022expertise}. For instance, in demonstration-guided reinforcement learning, expert data is added to improve sample efficiency \citep{rajeswaran2017learning} and to guide exploration \citep{nair2018overcoming}. Recent work has shown that including such expert information can help counter reward hacking in LLMs \citep{rita2024countering}. 

Consider the following medical example depicted in Figure~\ref{fig:motivating-example}. Patient 1 and Patient 2 consult the expert doctor and the student doctor about a condition they have. They have similar characteristics and essentially the same problem: a recurring morning headache that lasted a few days, but their exact phrasings can be different. Meanwhile, Patient 3 has different characteristics and a different condition to Patient 1 and 2. We think of the experienced doctor as representing a true policy and the student as a proxy policy. The two doctors both assign Patient 1 and 2 to the same group and Patient 3 to a different group, but the two doctors' recommendations for a given group are different, since the expert doctor can correctly recognised some easily misdiagnosed symptoms while the student doctor cannot.

We assume access to sparse prescriptions from an experienced doctor (ground truth) and plentiful prescriptions from less experienced student doctors (proxy). Even with ground truth data, If we naively learn a policy on the union of the dataset, we will learn a policy close to the proxy policy, as this data is more abundant. However, given the success of preference learning methods for LLMs, there is often useful information to extract from the prolific proxy data. A natural question is: \emph{When can proxy data ever provably improve preference learning?}

In this paper we outline a set of sufficient conditions on proxy feedback that, if satisfied, indicate that the proxy can provably improve the sample complexity of learning the ground truth policy. As not all proxies will satisfy these conditions, they can be used to guide a data collection process for a specific task. 
We show that as long as the collected proxy feedback shares certain properties with the true feedback, the sample complexity of learning with true preference data is provably improved by first training on large amounts of proxy preference data. The key idea behind this is that if the proxy and true policies share a certain structure, characterised in Condition~\ref{assp:shared-level-sets}-\ref{assp:functional-diff-is-lipschitz}, then it is possible to express the true policy as a low-dimensional adaptation of the proxy policy (Theorem~\ref{prop:d-dim-function-adjustment}). This relationship implies that certain parameters of the ground truth policy can be identified solely from proxy data, reducing the number of ground truth samples needed to learn the ground truth policy (Theorem~\ref{thm:covering-number-in-terms-of-dimension}, \ref{thm:sample-complexity-without-proxy}). This result immediately implies a parametrisation for LLMs that achieves improved sample complexity. 
Our contributions are:
\begin{itemize}[leftmargin=*]
    \item We characterise a set of sufficient conditions on proxy feedback that, if satisfied, the sample complexity of learning the true policy is reduced through learning on this proxy feedback.
    \item We show that if proxy feedback satisfies the sufficient conditions, it implies a specific model parametrisation and learning procedure to extract information from the proxies. We detail these and describe how one can adapt existing architectures to improve sample complexity.
\end{itemize}

\section{Preliminaries}
Consider the set of all prompts $\set{X}$ and completions $\set{Y}$, the elements of these sets are discrete sequences of tokens with arbitrary length, e.g. $\vec{x}=[x_1,\dotsc,x_{N_{\vec{x}}}]$. By considering an enumeration of all completions $\set{Y}$, the space of distributions $\P_{\set{Y}}$ is equivalent to the subset of positive and unit norm sequences in the sequence space $\ell^1$. We consider the set of policies, in the form of language models, which are maps $\set X \to \set{P_Y}$.


Starting from an initial policy $\pi_{\text{ref}}$\footnote{In practice, $\pi_{\text{ref}}$ is obtained from the supervised finetuning stage of language model training \citep{rafailov2023direct}.}, we want to find a target policy $\ctrue{\pidagger}: \set X \to \set {P_Y}$ which aligns with the preferences of an ideal actor in a given scenario. To learn this policy, we have preference data directly from the ideal actor, denoted $\ctrue{\set{D}^\dagger}$, as well as preference data from a proxy actor, $\cproxy{\tilde{\set{D}}}$. The central question we consider is: under what assumptions can $\cproxy{\tilde{\set{D}}}$ improve the sample complexity of learning $\ctrue{\pidagger}$?


\paragraph{Human preference feedback.} 
We aim to align $\pi_{\text{ref}}$ using preference data of the form $\{(\vec{x}, \vec{y}_w, \vec{y}_l)\}$, where $\vec{y}_w$ and $\vec{y}_l$ are candidate completions for prompt $\vec{x}$, and where $\vec{y}_w$ is preferred to $\vec{y}_l$. We assume these preferences are generated from a underlying scalar reward function $r(\vec{x}, \vec{y})$ according to
\citet{Bradley1952RankAO}:
\begin{gather}
    \vec{x}\sim p_\set{X};\quad
    \vec{y}_1,\vec{y}_2\underset{\text{i.i.d.}}{\sim}\pi_{\text{ref}}\callg{\blank}{\vec{x}};\\
    b \sim \operatorname{Bern}\bcall{\sigma\call{r\call{\vec{x},\vec{y}_1} - r\call{\vec{x},\vec{y}_2}}};\\
    (\vec{y}_w, \vec{y}_l) = \begin{cases*}
        (\vec{y}_1, \vec{y}_2) & if $b = 1$,\\
        (\vec{y}_2, \vec{y}_1) & if $b=0$.\\
    \end{cases*},
\end{gather}
where $\sigma(\blank)$ is the sigmoid logistic function.
We assume that $\vec y_1$ and $\vec y_2$ are sampled from $\pi_{\text{ref}}$ for simplicity, whereas in practice they can be sampled from other distributions over $\set{P_Y}$.
 In this model, higher relative rewards increase the chance of a completion being picked as the winner $\vec{y}_w$.
 
We assume that the true preference data $\ctrue{\set{D}^\dagger}$ and the proxy preference data $\cproxy{\tilde{\set{D}}}$ are generated according to distinct reward functions $\ctrue{r^\dagger}$ and $\cproxy{\tilde r}$. 

\paragraph{Bandit problem setting.} Given a data-generating process $G=\parens*{r, \pi_{\text{ref}}, p_{\set{x}}}$ with reward function $r$, a reference policy $\pi_{\text{ref}}$, and a distribution of prompts $p_{\set X}$, the optimal KL-regularised policy $\pi_G$ for the data generating process $G$ is the one that maximises the following optimisation objective:
\begin{align}
    \label{eq:bandit-objective}
    \arg\max_{\pi}
    \E[
        \vec{x}\sim p_\set{X},\; \vec{y}\sim\pi\callg{\blank}{\vec{x}}
    ]{r\call{\vec{x}, \vec{y}}} -
    \beta \KL{\pi\callg{\vec y}{\vec x}}{\pi_{\text{ref}}\callg{\vec y}{\vec x}},
\end{align}
where the regularisation parameter $\beta$ controls how close to the reference the optimum should be. Under this objective, the optimal policy is given by:
\begin{align}
    \label{eq:optimal-bandit}
    \pi_G\callg{\vec y}{\vec x} \propto
    \pi_{\text{ref}}\callg{\vec y}{\vec x}
    \exp\call{
        \frac{1}{\beta}r\call{\vec{x}, \vec{y}}
    }.
\end{align}

The target policy we aim to learn is thus denoted $\ctrue{\pi^\dagger}$ satisfying \eqref{eq:optimal-bandit} with respect to $\ctrue{G^\dagger} = \parens*{\ctrue{r^\dagger}, \pi_{\text{ref}}, \set{P_X}}$.

\paragraph{Direct preference optimisation (DPO) and implicit rewards.} 
Following \citet{rafailov2023direct}, by optimising the following objective:
\begin{align}
    \arg\max_{\pi}
    \E[
        (\vec{x}, \vec{y}_w, \vec{y}_l)\sim G
    ]{\log\sigma\call{
        \beta\log{\frac{\pi\callg{\vec{y}_w}{\vec{x}}}{\pi_{\text{ref}}\callg{\vec{y}_w}{\vec{x}}}} -
        \beta\log{\frac{\pi\callg{\vec{y}_l}{\vec{x}}}{\pi_{\text{ref}}\callg{\vec{y}_l}{\vec{x}}}}
    }},\label{eq:dpo-objective}
\end{align}
we recover the same optimal policy as described in Equation~\ref{eq:optimal-bandit} without directly using any reward function. Thus, DPO avoids the need for reward modelling.

We note that the policy implicitly defines a reward function via 
\begin{align}
r\call{\vec x, \vec y} &= \beta \log \frac{\pi\callg{\vec y_w }{ \vec x}}{\pi_{\text{ref}}\callg{\vec y_w}{\vec x}}
\end{align}

We can therefore define $\ctrue{\pidagger}, \cproxy{\pitilde}$ as the policies that (implicitly) define $\ctrue{r^\dagger}$ and $\cproxy{\tilde r}$, respectively.

\paragraph{True and proxy preference data} 
For most interesting tasks, sampling a dataset $\ctrue{\true{\set D}_{n^\dagger}} := \{\parens*{\vec{x}_i, \vec{y}_{w,i}, \vec{y}_{l,i}}\}_{i=1}^{\ctrue{n^\dagger}}$ from $\ctrue{G^\dagger}=(\ctrue{r^\dagger}, \pi_\text{ref},p_\set{x})$ can be costly, thus, the size $\ctrue{n^\dagger}$ of the dataset might not be large enough for adequate training. In these cases, it is common to use a much larger proxy dataset $\cproxy{\tilde{\set D}_{\tilde n}} := \{\parens*{\tilde{\vec{x}}_i, \tilde{\vec{y}}_{w,i}, \tilde{\vec{y}}_{l,i}}\}_{i=1}^{\cproxy{\tilde{n}}}$, where each data point is sampled i.i.d. from a proxy data-generating distribution $\cproxy{\tilde{G}}=(\cproxy{\tilde{r}}, \pi_\text{ref},p_\set{x})$, where $\cproxy{\tilde{G}}$ and $\ctrue{\true{G}}$ only differ in the reward function.

Nonetheless, if we do not have data from $\ctrue{\true{G}}$, then even if we have access to infinitely many data samples from $\cproxy{\tilde G}$, and even if it allows us to learn the perfect reward model $\cproxy{\tilde{r}}$\footnote{The technical condition for this to be possible is to be provided $\P_\set{X}$ and $\pi_{\text {ref}}$ have full support.}, we can at best only learn the optimal \emph{proxy} policy $\cproxy{\tilde{\pi}}$, which differs from the optimal \emph{true} policy in $\ctrue{\true{\pi}}$, \emph{by construction} due to the difference in rewards.

\section{Sufficient Conditions for Proxy Feedback}

Theory and survey papers point out the difficulty of alignment under mismatch between the {true reward function} and the one reflected by the human labelers \citep{skalse2022defining, casper2023open}. In other related fields such as vision, \citep{chi2022meta} shows the impossibility of leveraging pre-training data for unseen tasks unless some similarity between the two tasks is given. 

From these observations we can draw two conclusions. (i) We must have at least \emph{some} data from ${\color{true}\true{G}}$ to learn $\ctrue{\pidagger}$; this motivates the need for both $\ctrue{\set D^\dagger_n}$ and $\cproxy{\tilde{\set D}_{\tilde n}}$. (ii) In order for $\cproxy{\pitilde}$ to inform us something about $\ctrue{\pidagger}$, they must share some similarities. Thus, we ask the following research question: 

\emph{Under what sufficient conditions can we use $\cproxy{\tilde{\set D}}$ to learn $\ctrue{\pidagger}$ more efficiently than if we used $\ctrue{\set D^\dagger}$ alone? How much can we improve?}

With the following conditions, we show that, when we have access to large amounts of proxy data, $\ctrue{\pidagger}$ can be expressed as a low-dimension adaptation of $\cproxy{\pitilde}$, and hence the sample complexity of learning $\ctrue{\pidagger}$ is drastically improved, superexponential in the data manifold dimension.

While these conditions may appear strong, in practice, they can be helpful in guiding the design of the proxy data collection procedure. In particular, the conditions correspond to what expertise we might require proxy raters and true raters to share.

\begin{figure}[t!]
    \centering
    \includegraphics[width=1.\linewidth]{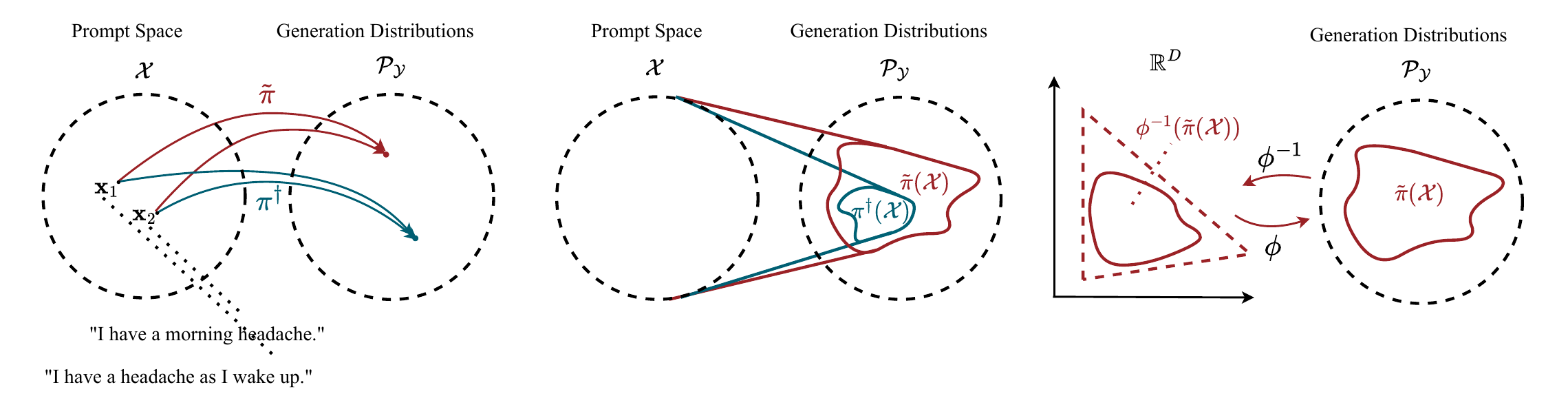}
    \caption{ Illustrations of conditions \ref{assp:shared-level-sets}-\ref{assp:image-in-convex-set}. Left, middle, right: Condition \ref{assp:shared-level-sets}, \ref{assp:shared-action-support}, \ref{assp:image-in-convex-set}, respectively.
    }
    \label{fig:conds-illustration}
\end{figure}

Our first condition says that given two distinct prompts, whenever they are mapped to the same response distribution under the true policy, so are they under the proxy policy; mathematically, the two policies share level sets (Figure~\ref{fig:conds-illustration}, left). 

\begin{assumption}[Shared level sets] \label{assp:shared-level-sets}
    Given $\vec{x}_1, \vec{x}_2 \in \set{X}$, we have that $\ctrue{\pidagger}\callg{\blank}{\vec{x}_1}=\ctrue{\pidagger}\callg{\blank}{\vec{x}_2}$ if and only if  $\cproxy{\pitilde}\callg{\blank}{\vec{x}_1} = \cproxy{\pitilde}\callg{\blank}{\vec{x}_2}$.
\end{assumption}

In the context of the running example (Figure~\ref{fig:motivating-example}), Condition~\ref{assp:shared-level-sets} corresponds to the experienced doctor and the student doctor classifying the symptoms of patients in the same way. We could reasonably expect this because comprehending the relevant details of a patient's query is part of the basic training for a doctor.
On the other hand, if the proxy preferences were sourced from generic crowd workers with no medical background, we would not expect this assumption to hold.

Our second condition says that the set of expert response distributions is contained within the set of proxy response distributions (Figure~\ref{fig:conds-illustration}, middle).

\begin{assumption}[Image inclusion]\label{assp:shared-action-support} All possible responses of the true policy are included in the proxy policy.
$\ctrue{\pidagger}\call{\set{X}} \subseteq \cproxy{\pitilde}\call{\set{X}}$.
\end{assumption}

In the context of the running example (Figure~\ref{fig:motivating-example}), Condition~\ref{assp:shared-action-support} says that the student doctor could, in principle, express any valid medical advice distribution, even if the student doctor may not know how to assign them with appropriate symptoms with high accuracy. 
This is again a reasonable assumption when the proxy feedback comes from a student doctor, but is less plausible if the proxy feedback comes from a rater without general knowledge of core concepts that underlie medical advice.


We also introduce two technical conditions.
Our first technical condition asserts that the image of $\cproxy{\pitilde}$ constitutes a smaller, easily characterized subset of the space of distributions over finite-length token sequences $\set{P_Y}$ (Figure~\ref{fig:conds-illustration}, right).
Roughly, this assumption is similar to a common condition underpinning many modern deep learning architectures: that data lie on a lower-dimensional manifold. 

\begin{assumption}[Finite-dimensional encoding of $\cproxy{\pitilde}\call{\set{X}}$]\label{assp:image-in-convex-set}
There exists an injective function $\phi\colon\set{V}\to\set{P_Y}$, where its domain $\set{V}\subset\Reals[D]$ is a bounded convex polytope with $D+1$ vertices, such that:
\begin{enumerate}
    \item Its image $\phi\call{\set{V}}$ contains the image of the policies: $\ctrue{\pidagger}\call{\set{X}} \subseteq \cproxy{\pitilde}\call{\set{X}} \subseteq \phi\call{\set{V}}$;
    \item It is $(L_\phi,L_{\phi^{-1}})$-bi-Lipschitz with its left inverse $\phi^{-1}\colon\set{P_Y}\to\set{V}$:
    $(1/L_{\phi^{-1}}) \norm{v_1-v_2}_p \leq d\call{\phi(v_1),\phi(v_2)} \leq L_\phi\norm{v_1-v_2}_p$, where $d$ is a metric on $\set{P_Y}$, discussed later in Section~\ref{sec:rates}.
\end{enumerate}
\end{assumption}

Note that the condition that $\set{V}$ is a convex polytope is benign, since we can extend any bi-Lipschitz function which bijects a compact subset of $\Reals[N]$ with $\cproxy{\pitilde}\call{\set X}$ to a bi-Lipschitz function from a bounded convex polytope with the same Lipschitz constants to a set containing $\cproxy{\pitilde}\call{\set X}$.



Condition~\ref{assp:image-in-convex-set} says that although the topological dimension of $\set{P_Y}$ can be extremely large, the image of $\cproxy{\pitilde}$ is identified with a Euclidean subset only of dimension $D \ll \operatorname{dim}[\set{P_Y}]$. 
To make this rigorous, we provide a proof that the  topological dimension of $\set{P_Y}$ can be as large as $\infty$; since any finite $n$-dimensional Euclidean space has topological dimension $n$, this shows that $\set {P_Y}$ is not a finite-dimensional Euclidean space.
\begin{proposition}[Topological dimension of ${\set{P_Y}}$ is $\infty$]\label{prop:dim-py-infinite}
    Let $\set{K} = \braces*{1, \cdots, k}$ denote a set of $k$ tokens. Let $\set Y$ be the set of all finite length token sequences whose tokens all come from $\set{K}$. Then $\set Y$ has a one-to-one identification with the natural numbers. Let $\set{P_Y}$ be the set of probability mass functions over $\set{Y}$, then the topological dimension $\operatorname{dim}\call{\set{P_Y}} = \infty$. If $\set{Y}$ is instead the set of token sequences with length $\leq l$, then $\operatorname{dim}\call{\set{P_Y}} = O\call{k^l}$. 
\end{proposition}
\begin{proof}
    Proof in Appendix~\ref{proof:dim-py-infinite}. 
\end{proof}
In practical LLM training regimes, a maximum sequence length is implemented, but the dimension of $\set{P_Y}$ grows exponentially with $l$.
In situations where the true and proxy policies generate responses from a small subset of all token sequences, it could be reasonable to expect Condition~\ref{assp:image-in-convex-set} to hold - for example, only a small subset of all token sequences form valid sentences, and an even smaller subset of those form valid medical advice, so we expect that for medical question-answering tasks the responses distributions can be encoded with fewer dimensions than for general question-answering tasks. The low-dimensional encoding can be viewed as some intermediate representation of the prompt that is sufficient for determining the response distribution. We can thus think of $\phi$ as a \emph{policy decoder}

Our final technical condition concerns the similarity between the policy functions $\ctrue{\pidagger}$ and $\cproxy{\pitilde}$. 

Since $\cproxy{\pitilde}$ and $\ctrue{\pidagger}$ share the same level sets, as supposed in Condition~\ref{assp:shared-level-sets}, it can be shown that $\ctrue{\pidagger} \circ {\cproxy{\pitilde ^{-1}}|_{\cproxy{\pitilde}\call{\set X}}}$ is a well-defined function, where it should be noted that ${\cproxy{\pitilde ^{-1}}|_{\cproxy{\pitilde}\call{\set X}}}\call{\vec{p}}$ maps a point $\vec p \in \cproxy{\pitilde}\call{\set X}$ to its pre-image under $\cproxy{\pitilde}$. The proof is provided in Appendix~\ref{proof:d-dim-function-adjustment}.

\begin{lemma}\label{lemma:pidagger-pitilde-inv-well-defined}
    Under Condition~\ref{assp:shared-level-sets}, {$\ctrue{\pidagger}\circ\cproxy{{\pitilde}^{-1}}|_{\cproxy{\pitilde}\call{\set{X}}}$ is a well-defined function.} 
\end{lemma}
Lemma~\ref{lemma:pidagger-pitilde-inv-well-defined} allows us to describe the `difference' between the proxy and true policy as a function: given $\vec p$, $\ctrue{\pidagger} \circ {\cproxy{\pitilde ^{-1}}|_{\cproxy{\pitilde}\call{\set X}}}$ maps \emph{all} input prompts which were assigned $\vec p$ by $\cproxy{\pitilde}$ to a distinct response distribution, say $\vec p'$, assigned by $\ctrue{\pidagger}$. This justifies learning an `adapter' function which reassigns $\vec p$ to the correct value $\vec p'$. Had Condition~\ref{assp:shared-level-sets} not held, then attempting to learn a function which assigns $\vec p$ to $\vec p'$ no longer makes sense since there could be some values of $\vec p$ for which the corresponding $\vec p'$ aren't unique. 

By Lemma~\ref{lemma:pidagger-pitilde-inv-well-defined}, we can impose a final technical condition which helps us quantify the difference between $\cproxy{\pitilde}$ and $\ctrue{\pidagger}$:
\begin{assumption}\label{assp:functional-diff-is-lipschitz}
    $\ctrue{\pidagger} \circ {\cproxy{\pitilde ^{-1}}|_{\cproxy{\pitilde}\call{\set X}}}$ is Lipschitz continuous.
\end{assumption}

Informally, this says that if two prompts are mapped to very similar completion distributions by $\cproxy{\pitilde}$, then they cannot get mapped to very different completion distributions by $\ctrue{\pidagger}$; for a unit difference of the former, the difference for the latter must not exceed $L$ for some positive scaler $L$. 

This condition applies to situations where the proxy rater is within reasonable ballpark from the true rater: for example, for some medicines the correct dosage can vary by a large amount depending on the patient; if the expert doctor prescribes a certain dosage, and the student doctor prescribes a dosage different but close to that, then the condition can be considered satisfied. The condition also generalises to a broader situation: if the expert doctor's \emph{change} in prescription, for instance, after observing some improvements in a patient, is similar to the \emph{change} in prescription of a student doctor, then the condition can also be considered satisfied. However, in situations of crowdsourcing human preference from places such as Amazon Mechanical Turk, we cannot consider this condition to be satisfied; likewise, if we suspect that some proxy raters are adversarial, then we also cannot expect it satisfied. 

\section{Adapting $\cproxy{\pitilde}$ to $\ctrue{\pidagger}$}

In this section, we derive an algorithm for learning $\ctrue{\pidagger}$ that leverages the structure in $\cproxy{\pitilde}$ specified in conditions~\ref{assp:shared-level-sets}-\ref{assp:functional-diff-is-lipschitz}.
The algorithm hinges on a decompositon of the policies $\pi$ into an encoder $\tau$, a linear layer $\Theta$, and a decoder $\phi$.
Importantly, we show that $\ctrue{\pidagger}$ can be expressed \emph{reusing} these components learned from proxy data generated by $\cproxy{\pitilde}$, with the addition of a \emph{low-dimensional adapter function between known spaces}. This reduction to learning an adapter function ultimately allows us to derive the sample complexity improvement (Theorem~\ref{thm:covering-number-in-terms-of-dimension}).

We outline the main steps of the derivation now:
\begin{enumerate}
    \item First, from Condition~\ref{assp:shared-action-support} and \ref{assp:image-in-convex-set}, we can show that both $\cproxy{\pitilde}$ and $\ctrue{\pidagger}$ map prompts into a common lower-dimensional space before decoding into response distributions. 
    \item Then, by Condition~\ref{assp:shared-level-sets} and \ref{assp:functional-diff-is-lipschitz}, it can be shown that $\cproxy{\pitilde}$ and $\ctrue{\pidagger}$ differ only by a Lipschitz continuous function mapping $\Delta^D \rightarrow \Delta^D$.
\end{enumerate}

\subsection{Factorising $\ctrue{\pidagger}$ and $\cproxy{\pitilde}$ through $\mathcal V$.}

Under Condition~\ref{assp:image-in-convex-set}, the proxy policy $\cproxy{\pitilde}$ factors through $\set V$: that is, one can view it as mapping from the space of prompts to the space of response distributions via some intermediate representation of the prompt (i.e., $\set V$) sufficient for determining the response distribution. Specifically, there is a bi-Lipschitz injective decoder $\phi$ from $\mathcal{V}$ to the image of the proxy policy $\cproxy{\pitilde}\call{\set X}$.

Condition~\ref{assp:shared-action-support} says that the image of $\ctrue{\pidagger}$ is included in the image of $\cproxy{\pitilde}$.

Therefore, we can view $\ctrue{\pidagger}$ as a function composition of some decoder ${\phi}: \mathcal{V} \to \cproxy{\pitilde} \call{\set X}$\footnote{Strictly, after $\ctrue{\phidagger}$ we still need to go through an inclusion to land in $\P_\set{Y}$, but to simplify notation we omit this technicality.} and some \emph{encoder} function $\set{X} \to \mathcal{V}$, such that
\begin{align}
    \ctrue{\pidagger} &: \set X \xrightarrow{{{\phi^{-1}} \circ \ctrue{\pidagger}}} \set V \xrightarrow{{\phi}} \cproxy{\pitilde}\call{\set X} 
\end{align}

Analogously, $\cproxy{\pitilde}$ can be viewed also as 
\begin{align}
    \cproxy{\pitilde} &: \set X \xrightarrow{{{\phi^{-1}} \circ \cproxy{\pitilde}}} \set V \xrightarrow{{\phi}} \cproxy{\pitilde}\call{\set X} 
\end{align}


Next, we show that $\ctrue{\pidagger}$ can be expressed by inserting a transformation into a function decomposition of $\cproxy{\pitilde}$. This transformation can be shown to be a Lipschitz map between two known $D$-dimensional spaces. We can thus think of this transformation as an `adapter' function.


\subsection{$\ctrue{\pidagger}$ and $\cproxy{\pitilde}$ differ by a function between $D$-simplices.} 

It is now possible to show that $\ctrue{\pidagger}$ and $\cproxy{\pitilde}$ differ only by a transformation on the representation space $\set V$. However, for the sample complexity arguments that follow, it is convenient to map $\set V$ to a $D$-simplex $\Delta^D$, then to show that $\ctrue{\pidagger}$ and $\cproxy{\pitilde}$ differ only by a transformation on $\Delta^D$.

To this end, since $\set V$ is a $D$-polytope with $D+1$ vertices, every point in $\set V$ can be expressed as a convex combination of the vertices. Then, it can be shown that $\cproxy{\pitilde}$ and $\ctrue{\pidagger}$ can be further factored through a $D$-simplex, $\Delta^D$. With this formalism in hand, we now state our result. 

\begin{theorem}\label{prop:d-dim-function-adjustment}
    We work under Conditions~\ref{assp:shared-level-sets}, \ref{assp:shared-action-support}, \ref{assp:image-in-convex-set} and \ref{assp:functional-diff-is-lipschitz}. For some $D$, there exists a Lipschitz invertible function $\cproxy{\tilde\phi}: \set V \to \cproxy{\pitilde}\call{\set X}$ satisfying Condition~\ref{assp:image-in-convex-set}, $\cproxy{\tilde\Theta} \in \Reals[N \times (D+1)]$ and $\cproxy{\tautilde^\circ}: \set X \to \Delta^D$ s.t. $\cproxy{\pitilde} =  \cproxy{\phitilde}\circ \cproxy{\tilde\Theta} \cproxy{\tautilde^\circ}$. 
    
    Moreover, for any $\parens*{\cproxy{\phitilde}, \cproxy{\tilde\Theta}, \cproxy{\tautilde^\circ}}$ such that $\cproxy{\pitilde} =  \cproxy{\phitilde}\circ \cproxy{\tilde\Theta} \cproxy{\tautilde^\circ}$, there exists a Lipschitz continuous function $\ctrue{\pibardagger}:\Delta^D \to \Delta^D$ s.t. $\ctrue{\pidagger} = \cproxy{\phitilde} \circ \cproxy{\tilde\Theta} \ctrue{\pibardagger} \circ \cproxy{\tautilde^\circ}$. 
\end{theorem}
\begin{proof}
   Proof in Appendix~\ref{proof:d-dim-function-adjustment}. 
\end{proof}



Theorem~\ref{prop:d-dim-function-adjustment} has two important implications for learning. 1. It establishes that there exists a decomposition of $\cproxy{\pitilde}$ with modules that can be reused to express $\ctrue{\pidagger}$. 2. It further establishes that \emph{for any} satisfactory decomposition, there exists an adapter $\ctrue{\pibar^{\dagger}}$. This suggests that in practice we can first find a suitable triplet $\parens*{\cproxy{\phitilde}, \cproxy{\tilde\Theta}, \cproxy{\tautilde^\circ}}$, then learn an adapter.

\subsection{Model Parametrisation and Learning}

We now sketch our learning algorithm. Theorem~\ref{prop:d-dim-function-adjustment} gives rise to a two-step procedure to learn the true policy $\ctrue{\pidagger}$, firstly we recover the functional components using a large proxy dataset $\cproxy{\tilde{\set D}_{\tilde n}}$ and then, secondly, we use a small true dataset $\ctrue{\set D^\dagger_{\ctrue{n^\dagger}}}$ to learn the low-dimensional adapter.

\paragraph{Stage 1} Based on Theorem~\ref{prop:d-dim-function-adjustment}, we model the proxy policy $\cproxy{\pitilde}$ with a parametric model composed of three functions: (i) $\cproxy{\tautildetheta^\circ}$, an embedding function from the prompts $\set X$ to the $D$-simplex $\Delta^D$, (ii) $\cproxy{\tilde\Theta}  \in \Reals[{D}\times{D}]$, a linear map from the simplex to a convex polytope $\set V$, and (iii) $\cproxy{\phitildetheta}$, an injective function from the latent space $\set V$ to a distribution of completions $\set{P_Y}$. Therefore, our model is expressed as:
\begin{align}
    \cproxy{\pitildetheta} &= \cproxy{\phitildetheta} \circ  \cproxy{\tilde\Theta} \cproxy{\tautildetheta^\circ} \label{eq:pi-tilde-theta-v0}.
\end{align}
Based on the DPO loss (Eq.~\ref{eq:dpo-objective}), $\cproxy{\phitildetheta}$, $\cproxy{\tilde\Theta}$, and $\cproxy{\tautildetheta^\circ}$ are learned using the empirical preference learning objective with the proxy dataset $\cproxy{\tilde{\set D}_{\tilde n}}$:
\begin{align}
    \ltilda_{{\color{proxy}\tilde n}}\call{\cproxy{\pitildetheta}} &=
    - \frac{1}{\tilde{n}}
    \sum_{i=1}^{\tilde{n}}
    \log\sigma\call{
        \beta\log
        \frac
            {\cproxy{\pitildetheta}\callg{\tilde{\vec{y}}_{w,i}}{\tilde{\vec{x}}_i}}
            {\pi_{\text{ref}}\callg{\tilde{\vec{y}}_{w,i}}{\tilde{\vec{x}}_i}}
        -\beta\log
        \frac
            {\cproxy{\pitildetheta}\callg{\tilde{\vec{y}}_{l,i}}{\tilde{\vec{x}}_i}}
            {\pi_{\text{ref}}\callg{\tilde{\vec{y}}_{l,i}}{\tilde{\vec{x}}_i}}
    }.
\end{align}
In the large sample limit, the optimal parametrised model $\cproxy{\phitildetheta^*} \circ \cproxy{\tilde\Theta^{*}}\cproxy{\tautildetheta^{\circ, *}}$ minimises the population proxy preference loss, and due to Theorem~\ref{prop:d-dim-function-adjustment}, we know that the optimal KL-regularised proxy policy $\cproxy{\pitilde} = \cproxy{\phitildetheta^*} \circ \cproxy{\tilde\Theta^{*}}\cproxy{\tautildetheta^{\circ, *}}$, thus, justifying our parametrization.

\paragraph{Stage 2} Following this, we then model the true policy $\ctrue{\pidagger}$ using the same pre-trained components from our model of the proxy policy $\cproxy{\pitildetheta}$ with the addition of a low-dimensional adapter function $\ctrue{\pibardaggertheta}$ which maps a latent representation in the simplex $\Delta^D$ to another $\Delta^D$ as follows:
\begin{align}
    \ctrue{\pidaggertheta} &= \cproxy{\phitildetheta} \circ \cproxy{\tilde\Theta} \ctrue{\pibardaggertheta} \circ \cproxy{\tautildetheta^\circ}
    \label{eq:pi-dagger-theta-v0}.
\end{align}

The adapter $\ctrue{\pibardaggertheta}$ is learned by optimization of the empirical preference learning objective with the true dataset $\ctrue{{\set D}^\dagger_{n^\dagger}}$, while keeping $\cproxy{\phitildetheta}$, $\cproxy{\tilde\Theta}$, and $\cproxy{\tautildetheta^\circ}$ fixed based on their previously optimised values:
\begin{align}
    \ctrue{\ldagger_{n^\dagger}}\call{\ctrue{\pibardaggertheta}} &=
    - \frac{1}{{\ctrue{n^\dagger}}}
    \sum_{i=1}^{{\ctrue{n^\dagger}}}
    \log\sigma\call{
        \beta\log
        \frac
            {\ctrue{\pibardaggertheta}\callg{{\vec{y}}_{w,i}}{{\vec{x}}_i}}
            {\pi_{\text{ref}}\callg{{\vec{y}}_{w,i}}{{\vec{x}}_i}}
        -\beta\log
        \frac
            {\ctrue{\pibardaggertheta}\callg{{\vec{y}}_{l,i}}{{\vec{x}}_i}}
            {\pi_{\text{ref}}\callg{{\vec{y}}_{l,i}}{{\vec{x}}_i}}
    }.
\end{align}

There can be more efficient algorithm which learn the triplet $\parens*{\cproxy{\phitilde, \tilde\Theta, \tautilde^\circ}}$ by using the proxy or true data simultaneously. But by splitting the learning into two stages, the first using only proxy data, and the second using only true preference data, we can make a direct sample complexity comparison between learning $\ctrue{\pidagger}$ from scratch, and learning the adapter $\ctrue{\pibardagger}$ in Stage 2, in terms of the size of the true preference data $\ctrue{{\set D}^\dagger_{n^\dagger}}$.

\section{Convergence Rates Analysis}\label{sec:rates}
To illustrate the benefit of learning $\ctrue{\pidagger}$ using the outlined algorithm, we analyse its soundness by showing the sample complexity improvement given that we have identified the true $\cproxy{\phitilde}$, $\cproxy{\tilde\Theta}$ and $\cproxy{\tautilde^\circ}$ from the proxy dataset in the first stage. This can be a reasonable approximation of the properties of the learning procedure in cases where the proxy dataset is much larger than the true dataset. To this end, we analyse the generalisation error bound for the second stage given access to true $\cproxy{\phitilde}$, $\cproxy{\tilde\Theta}$ and $\cproxy{\tautilde^\circ}$. Following the approaches of \citet{elesedy2022group} \citet[Exercise 3.31]{mohri2012foundations}, the generalisation error can be shown to be linear in the covering number of the hypothesis class. Our idea here is that the hypothesis class of $\pibar$ is made smaller by having knowledge of $\cproxy{\phitilde}$, $\cproxy{\tilde\Theta}$ and $\cproxy{\tautilde^\circ}$, hence the covering number is also smaller. In order to define covering numbers, we first need to define a notion of metric on all relevant spaces and the hypothesis classes we consider.

\subsection{Metrics and Hypothesis Classes}\label{subsec:metrics}
\paragraph{Metric on finite-dimensional vector spaces.}
For any finite-dimensional vector space, we use the $p$-norm-induced metric; for a simplex $\Delta^D$ we denote its metric by $d_\Delta$ and for any other finite dimensional space $\set U$ we use $d_{\set U}$. 

\paragraph{Metric on the prompt space.} The prompt space $\set X$, is a discrete and unstructured space, so we define a metric, $d_{\set X}$ on it based on some fixed embedding function $f$, which maps a prompt to a vector space with finite but high dimensions: $d_{\set X} \call{x, x'} = d_{f\call{\set X}} \call{f\call{x}, f\call{x'}} =  \norm*{f\call{x} - f\call{x'}}_p$.

For intuition we can think about $f$ as some general-purpose embedding, such as one got by retrieving some intermediate layer from a large model such as CLIP \cite{clip-radford}. Importantly, this metric is only relevant when considering the complexity of the hypothesis class when we learn the target policy without proxy data, which one expects to be large; this intuition is confirmed since a general-purpose embedding may not work well across all tasks, and can result in a large Lipschitz constant for the target function. 

\paragraph{Metric on a policy space.} Defining a metric on a space of policies is more involved, and we begin by relating a policy and the reward function under which it is the KL-regularised optimal policy. Given a policy $\pi: \set X \to \set{P_y}$, define the implicit reward as:
\begin{align}
    r_\pi\call{\vec{x},\vec{y}} &= \beta\log\frac{\pi\callg{\vec y}{\vec x}}{\pi_{\text{ref}}\callg{\vec y }{\vec x}}
\end{align}

Using $r_\pi$, we can define a metric on the set of policies, and we express in the next lemma. 

\begin{lemma} [$r_{\pi}$ helps define a metric] \label{lemma:rpi-helps-define-a-metric} 
For any $\pi, \pi': \set X \to \set{P_y}$ such that $\pi$ and $\pi'$ both satisfy $\norm*{r_\pi}_\infty \leq \infty$, define 
\begin{equation}
d_r\call{\pi,\pi'} = \|r_{\pi} - r_{\pi'}\|_\infty
\end{equation}
Then $d_r\call{\cdot,\cdot}$ defines a metric over the set of policies $
\braces*{
    \pi: \set{ x} \to \set{P_Y}\; \bigg| \; \sum_{\set Y} \pi\callg{y}{x} = 1
}
$.
\end{lemma}
\begin{proof}
    Proof in Appendix~\ref{proof:convergence-rate}.
\end{proof}

\paragraph{Metric on the completion distribution space.} We also need a notion of metric on the range of any policy function, $\set{P_Y}$. To this end, we define the following function $d_{\set{P_Y}}: \set{P_Y} \times \set{P_Y} \to \Reals \cup \braces*{\infty}$:
    \begin{align}
        d_{\set{P_Y}} \call{p, q} &:=  \norm*{\beta \log \frac{p\call{\cdot_y}}{q\call{\cdot_y}}}_\infty
    \end{align}
This construction is motivated by a resemblance with the construction of $r_\pi$. 

Now define the hypothesis class of $\ctrue{\pidagger}$ as:

\begin{definition}[Hypothesis class $\mathring{\Pi}$]\label{def:pi-ring}
    Let $\mathring \Pi$ be a set of functions $\pi: \set X \to \set{P_y} $ s.t. $\forall x \sum_y \pi\call{y|x} = 1 $, and $\pi$ satisfies $\norm*{r_\pi}_\infty \leq C$.
\end{definition}

So let $d_r$ be the metric on $\mathring\Pi$ and all its subsets. Later on, we will notice that the covering number of a hypothesis class depends on the smallest Lipschitz constant of the class. Let $\mathring\Pi\call{L}$ denote the subset of $\mathring\Pi$ such that every policy $\pi$ is $L$-Lipschitz wrt $d_{\set X}$ and $d_{\set P_Y}$.

Meanwhile, with knowledge of $\cproxy{\phitilde}, \cproxy{\tilde\Theta}, \cproxy{\tautilde^\circ}$, we define a subset of $\mathring\Pi$ with respect to a Lipschitz constant.

\begin{definition}[Hypothesis class fixing $\cproxy{\phitilde}, \cproxy{\tilde\Theta}, \cproxy{\tautilde^\circ}$ and Lipschitz-constant $L_{\pibar}$]
        Fix $\cproxy{\phitilde}, \; \cproxy{\tilde\Theta}, \; \cproxy{\tautilde^\circ}$. $\Pi\call{\cproxy{\phitilde}, \cproxy{\tilde\Theta}, \cproxy{\tautilde^\circ}, L_{\pibar}} \subseteq \mathring{\Pi}$ contains all $\pi \in \mathring\Pi$ which can be written as
    \begin{align}
        \pi\call{\cdot|x} &= \cproxy{\phitilde} \circ \cproxy{\tilde\Theta} \pibar \circ \cproxy{\tautilde^\circ} \call{x}
    \end{align}
    for some $L_{\pibar}$-Lipschitz $\pibar$, wrt $d_\Delta$.
\end{definition}

Now we are ready to state our main generalisation error bounds results.

\begin{theorem}[Bounding sample complexity in terms of dimension]\label{thm:covering-number-in-terms-of-dimension}
{
    We remain in the set up of Proposition~\ref{prop:covering-number-in-terms-of-domain-range}. The covering number of $\Pi$ is bounded above by a function of $D$:
    {\normalfont\begin{flalign}
        \mathrlap{\texttt{Cov}\call{\Pi, d_r, 3\kappa + 3L_{\phi}\norm{\cproxy{\tilde\Theta}}_p L_{\pibar}\delta}}
        &\notag\\
         &\leq \parens*{
            \frac{2 L_{\phi}\norm{\cproxy{\tilde\Theta}}_p\sqrt{D}}{\kappa}
        }^{
            D\parens*{
               \frac{2 \sqrt{D}}{\delta} 
            }^D
        }
    \end{flalign}}
    Set $\kappa = \frac{\epsilon}{48}$, we need 
    \begin{align}
    n\call{\epsilon, \omega}
    &= \Omega \call{
            \frac{D}{\epsilon^2}
            \parens*{
               \frac{
                    96L_{\phi}
                    \norm{\cproxy{\tilde\Theta}}_pL_{\pibar} 
               \sqrt{D}
               }{
               \epsilon 
            }
        }^D \log \parens*{
            \frac{96L_{\phi}\norm{\cproxy{\tilde\Theta}}_p \sqrt{D}}{\epsilon}
        }
        - \log \omega
    }
\end{align}
samples to generalise. That is, whenever $n' \geq n(\epsilon, \omega)$, we have 
\begin{align}
    P\bigg(\sup_{\pi \in \Pi} |R_G\call{\pi} - R_{\hat G_{n'}}\call{\pi} | \geq \epsilon \bigg) \leq \omega
\end{align}
}
\end{theorem}
\begin{proof}
    Proof in Appendix~\ref{proof:covering-number-in-terms-of-dimension}.
\end{proof}

\begin{theorem}[Bounding sample complexity of learning without proxy]\label{thm:sample-complexity-without-proxy}
    Let $D'$ be the dimension of a given embedding function $f$ of $\set X$.
{
    Let $\mathring\Pi\call{L_{\phi} \norm{\cproxy{\tilde\Theta}}_p L_{\pibar}}$ be the subset of $\mathring\Pi$ where $\pi$ is $L_{\phi} \norm{\cproxy{\tilde\Theta}}_p L_{\pibar}$-Lipschitz.
    Set $\kappa = \frac{\epsilon}{24}$, we need 
    \begin{align}
        \Omega \call{
                \frac{D'}{\epsilon^2}
                \parens*{
                   \frac{
                        48L_{\phi}
                        \norm{\cproxy{\tilde\Theta}}_p{L_{\pibar } }E'\call{p, D'}
                   \sqrt{D'}
                   }{
                   \epsilon 
                }
            }^{D'} \log \parens*{
                \frac{48L_{\phi}\norm{\cproxy{\tilde\Theta}}_p L_{\pibar} E'\call{p, D'} \sqrt{D'}}{\epsilon}
            }
            - \log \omega
        }
    \end{align}
samples to generalise, where $D' \gg D$ , and $E'\call{p, D'} \gg 1$. That is, whenever $n' \geq n(\epsilon, \omega)$, we have 
\begin{align}
    P\bigg(\sup_{\pi \in \mathring\Pi\call{L_{\phi} \norm{\cproxy{\tilde\Theta}}_p L_{\pibar}}} |R_G\call{\pi} - R_{\hat G_{n'}}\call{\pi} | \geq \epsilon \bigg) \leq \delta
\end{align}
}
\end{theorem}

\begin{proof}
    Proof in Appendix~\ref{proof:sample-complexity-without-proxy}.
\end{proof}

\paragraph{Discussion.} Theorem~\ref{thm:sample-complexity-without-proxy} says that if we learn $\ctrue{\pidagger}$ directly from expensive samples of $\ctrue{G^\dagger}$, then the sample complexity scales with $D'$, which is the dimension of the embedding space ; this can be extremely high dimensional. However if we parametrise $\ctrue{\pidagger}$ using $\cproxy{\phitilde}, \cproxy{\tilde\Theta}, \cproxy{\tautilde^\circ}$ which compose to be $\cproxy{\pitilde}$, and can be learned from cheap samples of $\tilde G$, then the number of \emph{expensive} samples we need from $\ctrue{G^\dagger}$ scale with $D$ which is assumed to be much lower-dimensional than $D'$.

\section{Related Work}

\paragraph{Reward hacking theory.} 
Initial work on the theory of reward hacking considered the setting where the proxy reward was a function of a subset of true reward features \citep{zhuang2020consequences}. This work demonstrates that optimising the proxy can lead to arbitrarily low true reward. Similarly, \citet{tien2022causal} give theoretical results for reward hacking when a learned reward uses nuisance variables that correlate with true causal variables. These ideas were extended to arbitrary MDPs by \cite{skalse2022defining} who define a proxy reward as hackable if it prefers policy $\pi_1$ over $\pi_2$ when the true reward has the opposite preference. Recent work has sought to develop scaling laws for reward hacking that describe how the true reward changes as the proxy reward is optimised \citep{gao2023scaling}. \citet{rafailov2024scaling} show similar over-optimisation patterns in DPO at higher KL-divergence budgets, even without an explicit reward model. In contrast to these works, our theoretical results suggest a new model parametrisation and training scheme which achieves improved sample complexity for learning the true policy; hence, our results are constructive.

\paragraph{Addressing reward hacking in LLMs.}
One of the classic examples of reward hacking in LLMs is their propensity for verbose responses that are not more helpful, often called the `length bias', or `length hacking' of LLMs \citep{singhal2023long}. To address this, \citet{singhal2023long} modified various aspects of PPO (increasing KL regularisation, omitting outputs beyond a certain length, and reward scaling), as well as the training data, with mixed success. \citet{chen2024odin} conducted a large-scale evaluation of the impact of hyperparameters and the above modifications on reward hacking. They further introduce a model that decorrelates preference predictions with length. \citet{miao2024mitigating} formulate reward modelling as optimising a variational information bottleneck and then use this to filter out less important features in latent space. \citet{huang2024correctingmythos} mitigates reward over-optimisation by replacing the KL regularisation with an alternative term which implicitly implements the principle of pessimism in the face of uncertainty.  
\citet{yang2024bayesianrewardmodelsllm} addresses a form of reward misalignment due to distribution shift of the prompts and responses seen in training versus test time. To the best of our knowledge, reward hacking due to a difference in the reward functions in training and test time is not discussed explicitly in existing work. More importantly, the impossibility of target policy recovery without \emph{some} data from the target reward, is not yet acknowledged.

\paragraph{RLHF with expert feedback.} 
Human feedback often varies in quality, and one key challenge is how to incorporate these different feedbacks into learning \citep{daniels2022expertise}. \citet{freedman2023active} formulate selecting which human to query for feedback as a bandit problem. \citet{yamagata2024relatively} uses the Boltzmann-rational model to account for varying levels of expertise. Our model parametrisation will leverage certain invariances between proxy and expert/true feedback that allows identification of the true policy as a low-dimensional adaptation of the proxy policy.

\section{Conclusion}

We study the problem of reward hacking due to distribution shifts. Specifically, an abundance of preference rankings is generated by a proxy reward function, different from the true reward function, which is costly to query.  We thus consider the setting where we have a large proxy dataset and a small true dataset; to the best of our knowledge, we are the first to consider this setting. We formulate conditions motivated by a real-world example, under which we prove that the optimal proxy policy can be decomposed into component functions shared with the optimal true policy, and that the true policy is only one low-dimensional adapter function away from the proxy policy given the shared component functions. We then observe that in the large sample limit of proxy data, one set of such component functions can be identified from minimising the preference loss. Leveraging this, we provide a characterisation of the sample complexity bound for learning the hypothesis class both with and without knowledge of the shared component functions; in particular, it is seen that under knowledge of the shared component functions the sample complexity bound is much lower than without such knowledge. 

As ongoing work, we are working on empirical evaluation of our theoretical findings, as well as relaxing some of the conditions we made. An updated version of the manuscript will be uploaded in the near future.


\subsubsection*{Acknowledgments}
YZ acknowledges support by the Engineering and Physical Sciences Research Council with grant number EP/S021566/1.

\bibliography{main,additional}
\bibliographystyle{iclr2025_conference}
\newpage
\appendix

\section{Proof of Proposition~\ref{prop:dim-py-infinite}}\label{proof:dim-py-infinite}
\textbf{Proposition~\ref{prop:dim-py-infinite}}
\emph{
    Let $\set{K} = \braces*{1, \cdots, k}$ denote a set of $k$ tokens. Let $\set Y$ be the set of all finite length token sequences whose tokens all come from $\set{K}$. Then $\set Y$ has a one-to-one identification with the natural numbers. Let $\set{P_Y}$ be the set of probability mass functions over $\set{Y}$, then the topological dimension $\operatorname{dim}\call{\set{P_Y}} = \infty$. If $\set{Y}$ is instead the set of token sequences with length $\leq l$, then $\operatorname{dim}\call{\set{P_Y}} = O\call{k^l}$.
}

\begin{proof}[Proof of Proposition~\ref{prop:dim-py-infinite}]

    We can identify $\set Y$ with the natural numbers as follows. 
    Let $l$ denote the length of a token sequence. Since for each token in the sequence there are $k$ options, there are $k^l$ distinct sequences of length $l$.
    
    We can define a bijective mapping from the set of $k^l$ sequences to the subset of natural numbers $\set{S}^l := \left\{\left(\sum_{i=1}^{l-1} k^i\right), \cdots, \left(\sum_{i=1}^l k^i\right) - 1 \right\}$ for $l\geq 2$ and $\set S^1 := \left\{0, \cdots, k-1\right\}$. This is possible since the number of elements in $\set S ^l$ is $\left(\sum_{i=1}^l k^i\right) - 1 - \left(\sum_{i=1}^{l-1} k^i\right) + 1 = k^l$. Denote one such mapping $f_l$. Then define $f: \set Y \to \mathbb N$:
    \begin{align}
        f(y) &= f_l(y) \text{ if length of $y = l$}
    \end{align}
    $f$ is well-defined because every $y$ has a unique length $l$. $f$ is invertible because $f_l$ is invertible for every $l$ and $\set S^l \cap \set S^{l'} = \emptyset$ and $\bigcup_{l=1}^\infty \set S^l = \mathbb N$.

    Thus, $\set{P_Y}$ is the set of probability mass functions whose sample space is $\cong \mathbb N$. That is, an element $P_Y \in \set{P_Y}$ is an infinite positive sequence which sums to $1$. 
    
    Let $\Delta^d$ be the $d$-dimensional simplex. Note that $\mathcal{P_Y} = \bigcup_{d=1}^\infty \Delta^d$, where $\Delta^d$ is viewed as a subset of $\Delta^{d+1}$ via the inclusion $\Delta^d \hookrightarrow \Delta^{d+1}:\;(p_1, \cdots, p_{d+1}) \mapsto (p_1, \cdots, p_{d+1}, 0, \cdots)$. 
    
    For each $d$, the topological dimension of the $d$-simplex $\Delta^d$ is $d$ \footnote{Theorem 5, https://personal.colby.edu/~sataylor/teaching/F14/MA331/TopologicalDimension.pdf}; this is to say, $d$ is the smallest number such that every open cover of $\Delta^d$ has an open refinement of order $d+1$. Therefore, the smallest number $n$ such that every open cover of $\bigcup_{d=1}^D \Delta^d$ has an open refinement of order $n+1$ is $D$. Hence, there is no finite $N$ such that the any open cover of $\mathcal{P_Y} := \bigcup_{d=1}^\infty \Delta^d$ has an open refinement with order $N+1$. Therefore the topological dimension of $\mathcal{P_Y}$ is $\infty$. 

    When the maximum sequence length is $l$ the cardinality of $\set{Y}$ is finite and equal to $\sum_{i = 1}^l  k^i = \frac{k(1-k^l)}{1-k}$. $\set{P_Y}$ thus contains the set of positive sequences of length $\frac{k(1-k^l)}{1-k}$ which sum to 1, so $\set{P_Y}$ is a $\frac{k(1-k^l)}{1-k} - 1$-dimensional simplex, and therefore the topological dimension of $\set{P_Y}$ is $\frac{k(1-k^l)}{1-k} - 1 = O\call{k^l}$.
    
\end{proof}

\section{Proof of Theorem~\ref{prop:d-dim-function-adjustment}} \label{proof:d-dim-function-adjustment}

\textbf{Lemma~\ref{lemma:pidagger-pitilde-inv-well-defined}}
\emph{
    Under Condition~\ref{assp:shared-level-sets}, {$\ctrue{\pidagger}\circ{\cproxy{\pitilde}^{-1}}|_{\cproxy{\pitilde}\call{\set{X}}}$ is a well-defined function.} 
}

\begin{proof}\;
\paragraph{Inverse of non-injective functions.} In general, unless a function is \emph{injective}
\footnote{Injective essentially means one-to-one. Formally, a function $f$ is injective if $f\call{\vec{x}_1} \neq f\call{\vec{x}_2}$ whenever $\vec{x}_1 \neq \vec{x}_2$.}
,
its inverse is not a function, but only a set map. For instance, since $\cproxy{\pitilde}$ is many-to-one, so \emph{not} injective, $\cproxy{\pitilde^{-1}}$ would take an element from $\P_\set{Y}$ and return a \emph{subset} of $\set X$, rather than a single element. Let
$\cproxy{{\pitilde}^{-1}}|_{\cproxy{\pitilde}\call{\set{X}}}$ denote the inverse of $\cproxy{\pitilde^{-1}}$ restricted to its image $\cproxy{\pitilde}\call{\set{X}}$. 

\textbf{$\ctrue{\pidagger}\circ\cproxy{{\pitilde}^{-1}}|_{\cproxy{\pitilde}\call{\set{X}}}$ is a well-defined function.} 
It follows that $\ctrue{\pidagger}\circ\cproxy{{\pitilde}^{-1}}|_{\cproxy{\pitilde}\call{\set{X}}}$ 
takes a point $P_Y$ in the image of $\cproxy{\pitilde}$, $\cproxy{\pitilde}\call{\set X}$, map it to its preimage $\cproxy{\pitilde^{-1}}\call{P_Y}$, and map all points in the preimage $\cproxy{\pitilde^{-1}}\call{P_Y}$ through $\ctrue{\pidagger}$ to $\ctrue{\pidagger}\call{\cproxy{\pitilde^{-1}}\call{P_Y}}$. For any two points $\vec x_1, \; \vec x_2 \in \cproxy{\pitilde^{-1}}\call{P_Y}$, we have $\cproxy{\pitilde}\call{\vec x_1} = \cproxy{\pitilde}\call{\vec x_2}$, and then Condition~\ref{assp:shared-level-sets} implies $\ctrue{\pidagger}\call{\vec x_1} = \ctrue{\pidagger}\call{\vec x_2}$. Therefore, for any $P_Y \in \set{P_Y}$, $\ctrue{\pidagger}\call{\cproxy{\pitilde^{-1}}\call{P_Y}}$ is a set containing exactly one element, so $\ctrue{\pidagger}\circ {\cproxy{\pitilde^{-1}}}|_{\cproxy{\pitilde}\call{\set X}}$ is a well-defined function. 
\end{proof}

\textbf{Theorem~\ref{prop:d-dim-function-adjustment}}
\emph{
    We work under Assumptions~\ref{assp:shared-level-sets}, \ref{assp:shared-action-support}, \ref{assp:image-in-convex-set} and \ref{assp:functional-diff-is-lipschitz}. For some $D$, there exists a Lipschitz invertible function $\cproxy{\tilde\phi}: \set V \to \cproxy{\pitilde}\call{\set X}$, $\cproxy{\tilde\Theta} \in \Reals[N \times (D+1)]$ and $\cproxy{\tautilde^\circ}: \set X \to \Delta^D$ s.t. $\cproxy{\pitilde} =  \cproxy{\phitilde}\circ \cproxy{\tilde\Theta} \cproxy{\tautilde^\circ}$, and there is a Lipschitz continuous function $\ctrue{\pibardagger}:\Delta^D \to \Delta^D$ s.t. $\ctrue{\pidagger} = \cproxy{\phitilde} \circ \cproxy{\tilde\Theta} \ctrue{\pibardagger} \circ \cproxy{\tautilde}$. 
}

\begin{proof}[Proof of Proposition~\ref{prop:d-dim-function-adjustment}]
    \textbf{Step 1. Show that there exists $\cproxy{\tilde\Theta}$, $\cproxy{\tautilde^\circ}$ and Lipschitz $\cproxy{\tilde\phi}$ such that $\cproxy{\pitilde} = \cproxy{\phitilde} \circ \cproxy{\tilde\Theta} \cproxy{\tautilde^\circ}$.}

    By Condition~\ref{assp:image-in-convex-set}, there is some invertible $L_{\phi}$-Lipschitz function $\cproxy{\phitilde}: \set{V} \to \cproxy{\pitilde}\call{\set X}$, where $\mathcal{V}$ is some convex polygon. Therefore, there is a finite set $\set{V}_{D+1} = \{\vec v_d\}_{d=1}^{D+1}$ such that every $\vec v \in \set V$ can be expressed as $\vec v = \sum_{d=1}^{D+1} p_d \vec v_d$ for some $\vec p \in \Delta^D$. Let $\cproxy{\tilde\Theta} \in \Reals[N\times (D+1)]$ be the matrix such that its $d$-th column, $\cproxy{\tilde\Theta}_{:, d}$, is equal to $\vec v_d$. Then every $\vec v \in \set V$ can be written as $\vec v := \cproxy{\tilde\Theta} \vec p$ for some $\vec p \in \Delta^D$. 

    Since $\cproxy{\phitilde^{-1}} \circ \cproxy{\pitilde}$ is a function $\set X \to \set V$, then for every $x$, $\cproxy{\phitilde^{-1}} \circ \cproxy{\pitilde}\call{x}$ is in $\set V$. Therefore, there exists some $\vec p_x \in \Delta^D$ such that
    \begin{align}
        \cproxy{\phitilde^{-1}} \circ \cproxy{\pitilde} \call{x} &= \cproxy{\tilde\Theta} \vec p_x
    \end{align}

    Let $\cproxy{\tautilde^\circ}: \set X \to \Delta^D$ be s.t. 
    \begin{align}
        \cproxy{\tautilde^\circ} \call{x} &= \vec p_x
    \end{align}
    then 
    \begin{align}
        \cproxy{\phitilde^{-1}}\circ \cproxy{\pitilde}\call{x} &= \cproxy{\tilde\Theta} \cproxy{\tautilde^\circ}\call{x} \\
        \cproxy{\pitilde}\call{x} &= \cproxy{\tilde\phi} \circ \cproxy{\tilde\Theta} \cproxy{\tautilde^\circ}\call{x}
    \end{align}

    \paragraph{Step 2. Let $\cproxy{\tautilde}\call{x} = \cproxy{\tilde\Theta}\cproxy{\tautilde^\circ}\call{x}$. We show that under the shared-level-sets assumption, $\ctrue{\pidagger}\circ \cproxy{\tautilde^{-1}}|_{\cproxy{\tautilde}\call{\set X}}$ is well-defined.}
    We have the following equalities: 
    \begin{align}
        \cproxy{\pitilde^{-1}}|_{\cproxy{\pitilde}\call{\set X}} &= \parens*{
            \cproxy{\phitilde}
            \circ 
            \cproxy{\tautilde}
        }^{-1}|_{
            \cproxy{\pitilde}\call{\set X}
        } \\
        &= {
            \cproxy{\tautilde}
        }^{-1}|_{
            \set V
        } 
        \circ
        \cproxy{\phitilde^{-1}} \\[0.5em]
        \cproxy{\pitilde^{-1}}|_{\cproxy{\pitilde}\call{\set X}} \circ \cproxy{\phitilde}
        &= {
            \cproxy{\tautilde}
        }^{-1}|_{
            \set V
        } \\
        &= {
            \cproxy{\tautilde}
        }^{-1}|_{
            \cproxy{\phitilde^{-1}} \call{\cproxy{\pitilde}\call{\set X}}
        } \\
        &= {
            \cproxy{\tautilde}
        }^{-1}|_{
            \cproxy{\tautilde}\call{\set X}
        }
    \shortintertext{Therefore,}
        \cproxy{\pitilde^{-1}}|_{\cproxy{\pitilde}\call{\set X}} \circ \cproxy{\phitilde}
        &= {
            \cproxy{\tautilde}
        }^{-1}|_{
            \cproxy{\tautilde}\call{\set X}
        } \\
        \ctrue{\pidagger} \circ \cproxy{\pitilde^{-1}}|_{\cproxy{\pitilde}\call{\set X}} \circ \cproxy{\phitilde}
        &= \ctrue{\pidagger} \circ {
            \cproxy{\tautilde}
        }^{-1}|_{
            \cproxy{\tautilde}\call{\set X}
        } \\
    \end{align}
    By Condition~\ref{assp:shared-level-sets} and Lemma~\ref{lemma:pidagger-pitilde-inv-well-defined}, $\ctrue{\pidagger}\circ \cproxy{\pitilde^{-1}}|_{\cproxy{\pitilde}\call{\set X}}$ is well-defined. 
    Since $\ctrue{\pidagger}\circ \cproxy{\pitilde^{-1}}|_{\cproxy{\pitilde}\call{\set X}}$ is well-defined, so is $\ctrue{\pidagger} \circ {
            \cproxy{\tautilde}
        }^{-1}|_{
            \cproxy{\tautilde}\call{\set X}
        } $.

    \paragraph{Step 3. Show that under Assumptions~\ref{assp:shared-action-support}, \ref{assp:image-in-convex-set} and \ref{assp:functional-diff-is-lipschitz}, $\ctrue{\pidagger}$ can be decomposed as $\cproxy{\phitilde}\circ \psi \circ \cproxy{\tautilde}$ for some Lipschitz function $\psi:\cproxy{\tautilde}\call{\set X} \to \cproxy{\phitilde^{-1}} \call{\ctrue{\pidagger}\call{\set X}}$.}
    Note that 1. $\cproxy{\phitilde}$ is invertible restricted to its image, 2. by Condition~\ref{assp:shared-action-support} and \ref{assp:image-in-convex-set} $\cproxy{\phitilde^{-1}}$ is defined on the image of $\ctrue{\pidagger}$, and 3. $\ctrue{\pidagger}\circ\cproxy{\tautilde^{-1}}|_{\cproxy{\tautilde} \call{\set X}}$ is well-defined. Therefore, we can factor $\ctrue{\pidagger}$ as 
    \begin{align}
        \ctrue{\pidagger} &= \cproxy{\phitilde}\circ \cproxy{\phitilde^{-1}} \circ \ctrue{\pidagger} \circ \cproxy{\tautilde^{-1}}|_{\cproxy{\tautilde} \call{\set X}} \circ \cproxy{\tautilde}
    \end{align}

    Therefore, define:
    \begin{align}
        \psi:\cproxy{\tautilde}\call{\set X} &\to \cproxy{\phitilde^{-1}} \call{\ctrue{\pidagger}\call{\set X}} \subset \cproxy{\tautilde}\call{X} \\
        \psi:={}& {\cproxy{\phitilde^{-1}} \circ \ctrue{\pidagger} \circ \cproxy{\tautilde^{-1}}|_{\cproxy{\tautilde} \call{\set X}}} \\
        ={}& {\cproxy{\phitilde^{-1}} \circ \ctrue{\pidagger} \circ \cproxy{\pitilde^{-1}}|_{\cproxy{\pitilde} \call{\set X}}} \circ \cproxy{\phitilde}
    \end{align}
    is a composition of Lipschitz functions (by Assumptions~\ref{assp:image-in-convex-set} and \ref{assp:functional-diff-is-lipschitz}) so is Lipschitz.

    \paragraph{Step 4. Finally show the assertion, that $\ctrue{\pidagger} = \cproxy{\phitilde} \circ \cproxy{\tilde\Theta} \ctrue{\pibardagger} \circ \cproxy{\tautilde^\circ}$ for some Lipschitz $\ctrue{\pibardagger}: \Delta^D \to \Delta^D$.} 
    Substituting in $\cproxy{\tautilde}\call{x} = \cproxy{\tilde\Theta} \cproxy{\tautilde^\circ}\call{x}$, we obtain:
    \begin{align}
        \ctrue{\pidagger}\call{x} &= \cproxy{\phitilde} \circ \psi \circ \cproxy{\tilde\Theta} \cproxy{\tautilde^\circ}\call{x}
    \end{align}
    Let $\vec p_x = \tau^\circ\call{x} \in \Delta^D$. We want to show that there is a Lipschitz continuous function $\ctrue{\pibardagger}: \Delta^D \to \Delta^D$ such that
    \begin{align}
        \psi\parens[\big]{\cproxy{\tilde\Theta} \vec p_x} &= \cproxy{\tilde\Theta}\ctrue{\pibardagger}\parens[\big]{\vec p_x}
    \end{align}
    For a given $\vec p_x$, we can try to solve the linear system in terms of $\ctrue{\pibardagger}\call{\vec p}$. We know that it must be an element of the set:
    \begin{align}
        \cproxy{\tilde\Theta^+}\psi\parens[\big]{\cproxy{\tilde\Theta} \vec p_x} + \operatorname{Ker}\parens[\big]{\cproxy{\tilde\Theta}},
    \end{align}
    where $\cproxy{\tilde\Theta^+}$ denotes the pseudoinverse.
    
    Take the intersection between this set and $\Delta^D$; the intersection is non-empty because $\psi$ lands in $\set V$. Now we describe a procedure to choose a point in this intersection that is Lipschitz continuous wrt $\vec p_x$: we let $\ctrue{\pibardagger}$ map $\vec p_x$ to the centroid of the intersection between $\Delta^D$ and $\cproxy{\tilde\Theta^+}\psi\parens[\big]{\cproxy{\tilde\Theta} \vec p_x} + \operatorname{Ker}\parens[\big]{\cproxy{\tilde\Theta}}$. The intersection is one of two convex sets, so it is convex; so the centroid lie in this set.

    We now proceed to show that $\ctrue{\pibardagger}$ is Lipschitz continuous, in two steps. 1. First we show that the centroid of the intersection is a smooth function of the location of its vertices. 2. Then we show that the location of the vertices is a piecewise smooth function of $\vec p_x$.
    \paragraph{We show the centroid of the intersection is a generically smooth function of the location of its vertices.}
    Note that the intersection of $\Delta^D$ and $\cproxy{\tilde\Theta} \psi\parens[\big]{\cproxy{\tilde\Theta} \vec p_x} + \operatorname{Ker}\parens[\big]{\cproxy{\tilde\Theta}}$ is a convex high-dimensional polyhedron, denote it $\set S \call{\vec p_x}$. 

    The centroid of a convex high-dimensional polyhedron can be computed as follows: every convex polyhedron admits a triangulation. Let the triangulation of $\set S \call{\vec p_x}$, denote it by $\operatorname{T}\call{\set S\call{\vec p_x}}$. For the $i$th simplex in $\operatorname{T}\call{\set S\call{\vec p_x}}$, take its vertices $\braces*{\vec v_{i0}, \cdots, \vec x_{in}}$ where $n$ is the dimension of the polyhedron. The centroid of the simplex is given by $\operatorname{C}\call{i} = \frac{\vec v_{i0}+\cdots + \vec v_{in}}{n+1}$, and the volume $\operatorname{Vol}\call{i}$ is given by $\frac{1}{n!} \abs*{\begin{pmatrix}
        \vec v_0 & \cdots & \vec v_n \\
        1 & \cdots & 1
    \end{pmatrix}}$. 
    The centroid of $\set S \call{\vec p_x}$, denoted $\operatorname{C}\call{\set S \call{\vec p_x}}$ is given by 
    $\sum_i \operatorname{C}\call{i} \operatorname{Vol}\call{i}$; since the determinant function is a polynomial in the matrix entries, this is a vector field where each entry is a polynomial. Moreover, every vertex in the triangulation but is not on $\set S \call{\vec p_x}$ is in the interior of $\set S \call{\vec p_x}$. There is some $\epsilon$ small enough such that we can draw an $\epsilon$-ball around each such vertex such that the closure of the balls are all disjoint and still lie in the polyhedron. So, perturb move each vertex to a point on the boundary of its ball; this gives a new triangulation, but the centroid is not changed. Since we can do it for any $\epsilon' \leq \epsilon$, $\operatorname{C}\call{\set S \call{\vec p_x}}$ is constant in the interior vertices. Therefore, $\operatorname{C}\call{\set S \call{\vec p_x}}$ is a polynomial of its vertices. 
    
    The limiting case is when two vertices overlap; in this case, at least one element in the triangulation will collapse onto a face, which has volume zero, so its centroid will not contribute to the calculation of $\operatorname{C}\call{\set S \call{\vec p_x}}$ through the formula. Therefore, the centroid of a polyhedron is a polynomial of its vertices, including when two or more vertices overlap.

    \paragraph{We show that the location of the vertices is a piecewise-smooth function of $\vec p_x$.} 
    Note that the set of points in $\set S \call{\vec p_x}$ is described as follows: Suppose $\operatorname{dim}\call{\set S \call{\vec p}} = J \leq D$, and choose an orthonormal basis in $\Reals[D+1]$ whose span contains the direction vectors in $\set S \call{\vec p}$:
    \begin{align}
        \vec b_1, \cdots, \vec b_J
    \end{align}
    Extend this to an orthonormal basis whose span contains $\Delta^D$:
    \begin{align}
        \vec b_1, \cdots, \vec b_J, \vec b_{J+1}, \cdots, \vec b_{D}
    \end{align}
    And finally extend this to $\Reals[D]$:
    \begin{align}
        \vec b_1, \cdots, \vec b_J, \vec b_{J+1}, \cdots, \vec b_{D}, \vec b_{D+1}
    \end{align}
    So we can express 
    \begin{align}
        \set S\call{\vec p_x} =
        \bigg\{ 
            \vec s \in \Reals[D+1] \; \bigg| \; 
            s_i \geq 0, 
            \sum_i s_i = 1,
            \mat{A} \vec s =
            \mat{A} \parens*{\cproxy{\tilde\Theta^+ }\psi\parens[\big]{\cproxy{\tilde\Theta} \vec p}}
        \bigg\},
    \end{align}
    where $\mat{A} = \parens[\big]{\mat I_{D+1} - \mat B_J \mat B_J^\top }$, $\mat B$ is the matrix whose rows are the $D+1$ basis vectors, and $\mat B_J$ is the one taking its first $J$ rows.

    Note that $\mat I_{D+1} - \mat B_J \mat B_J^\top = \mat B_{-J} \mat B_{-J}^\top $ where $\mat B_{-J}$ is the matrix containing $\vec b_{J+1}, \cdots, \vec b_{D+1}$ as rows. But $\vec 1$ is orthogonal to the row space of $\mat B_J^\top$, so it is contained in the row space of $\mat B_{-J}^\top$, and hence $\mat B_{-J}\mat B_{-J}^\top$. Therefore, we can remove $\sum_i s_i = 1$ from the set of conditions. 

    Therefore, the set of conditions contains $D+1 - J$ linearly independent conditions and $D+1$ inequalities.
    
    An extrema, i.e. a vertex, is the solution of $D+1$ linearly independent equations where all $D+1 - J$ linearly independent equality constraints are included, together with $J$ equations from saturating the inequality constraints. Since $\mat I_{D+1} - \mat B_J \mat B_J^\top $ has rank $D+1 - J$, there is a subset of $D+1-J$ rows, call the new matrix constructed from these rows $\bar {\mat{B}}_{D+1-J}$. Select $J$ vectors from the standard basis which are linearly independent of the rows of $\bar {\mat{B}}_{D+1-J}$, and stack them into an invertible matrix $\mat C$. Then any vertex is a solution of one such equation
    \begin{align}
        \vec v^* &= \mat C^{-1} \begin{pmatrix}
            \bar{\mat{B}}_{D+1 -J}\parens*{\cproxy{\tilde\Theta^+} \psi\call{\cproxy{\tilde\Theta} \vec p_x}}\\
            \vec 0
        \end{pmatrix}
    \end{align}
    provided it still satisfies the remaining inequality constraints. Here it is clear that any $\vec v^*$ varies smoothly with $\vec p_x$. When all vertices of $ \set S\call{\vec p_x} $ are in the interior of a $1$-dimensional face of $\Delta^D$, they vary locally smoothly with $\vec p_x$. 
    Since the centroid of $\set S \call{\vec p_x}$ varies smoothly with its vertices, whenever its vertices vary smoothly with $\vec p_x$, the centroid also vary locally smoothly with $\vec p_x$. The only non-differentiability happens when one vertex moves out of $\Delta^D$ and another moves in. But around those points of $\vec p_x$ the centroid is still continuous wrt $\vec p_x$ , so $\ctrue{\pibardagger}$ is piecewise differentiable function on a compact domain, and therefore is Lipschitz continuous.
\end{proof}

\section{Convergence rates proofs} \label{proof:convergence-rate}

\textbf{Lemma~\ref{lemma:rpi-helps-define-a-metric}. }($r_{\pi}$ helps define a metric) \emph{
Define $r_{\pi}\call{\vec{x}, \vec{y}} = \beta \log \frac{\pi\callg{\vec y}{\vec x}}{\pi_{\text{ref}}\callg{\vec y}{\vec x}}$ for some fixed constant $\beta > 0$, and let 
\begin{equation}
d_r\call{\pi,\pi'} = \|r_{\pi} - r_{\pi'}\|_\infty
\end{equation}
Then $d_r\call{\cdot,\cdot}$ defines a metric over any set $\Pi$ of functions  
$\mathcal X \times \mathcal Y \rightarrow [0, 1]$ s.t. $\forall \vec{x} \sum_\set{y} \pi\callg{\vec y}{\vec x} = 1$, 
and satisfies $\abs*{r_\pi\call{\vec{x},\vec{y}}} \leq C$ over $\mathcal X$ and $\mathcal Y$.
}
\begin{proof}[Proof of Lemma~\ref{lemma:rpi-helps-define-a-metric}]
    $d_r$ is well-defined on $\Pi$ since for any $\pi, \pi' \in \Pi$, $d_r\call{\pi, \pi'} \leq \norm{r_\pi}_\infty + \norm{r_{\pi '}}_\infty \leq 2C < \infty$. 
    
    We can verify that $d$ is a metric on $\Pi$. Clearly, symmetry and positivity holds, and $d\call{\pi,\pi '} = 0 \iff \pi = \pi'$, 
so we just need to check triangle inequality. Fix $\pi'' \in \Pi$, 

\begin{align}
d_r\call{\pi,\pi''} + d_r\call{\pi',\pi''}
&= \|r_{\pi} - r_{\pi''}\|_\infty + \|r_{\pi'} - r_{\pi''}\|_\infty\\
&\geq\|r_{\pi} - r_{\pi''} - r_{\pi'} + r_{\pi''}\|_\infty \\
&=\|r_{\pi} - r_{\pi'}\|_\infty \\
&= d_r\call{\pi,\pi'}
\end{align}
\end{proof}

\begin{proposition}[Concentration bound]\label{prop:concentration-bound}
Let $G$ be a measure on $(X, Y_w, Y_l)$ and for any $\pi\in \Pi \subseteq \mathring\Pi $ (Def.~\ref{def:pi-ring}) let
\begin{equation}
R_{G}\call{\pi} = \mathbb{E}_{G}\left[\log \sigma \left(\beta \log \frac{\pi\call{Y_w|X}}{\pi_{\text{ref}}\call{Y_w|X}} - \beta \log \frac{\pi\call{Y_l|X}}{\pi_{\text{ref}}\call{Y_l|X}}  \right)\right]
\end{equation}
be the preference loss. Further, let $(X_i, Y_{w, i}, Y_{l, i})_{i=1}^n$ be i.i.d. samples from $G$, and let $\hat G_n$ denote the empirical measure given by the samples, then 
\begin{equation}
P\bigg(\sup_{\pi \in \Pi} |R_G\call{\pi} - R_{\hat G_n}\call{\pi} | \geq \epsilon \bigg) \leq 2 \inf_{\alpha \in \parens*{0,1}} \mathrm{Cov}\call{\Pi, d_r(\cdot, \cdot), \frac{\alpha \epsilon}{4}} e^{-\frac{2\call{1-\alpha}^2n\epsilon^2}{4C^2}}
\end{equation}
where $d_r(\pi, \pi') = \|r_\pi - r_{\pi'}\|_\infty$.
\end{proposition}

\begin{proof}[Proof of Proposition~\ref{prop:concentration-bound}.]
Adapted from \citet{elesedy2022group}.

Fix $\pi, \pi' \in \Pi$.

\begin{align}
\abs*{R_G\call{\pi} - R_G\call{\pi'}}
={}&  \bigg|\mathbb{E}_{G}\bigg[
    \log\sigma\parens*{
        \beta \log \frac{\pi\call{Y_w|X}}{\pi_{\text{ref}}\call{Y_w|X}}-\beta \log \frac{\pi\call{Y_l|X}}{\pi_{\text{ref}}\call{Y_l|X}}
    }
\nonumber\\&\;\;\;\;\;\;\;\;
    - \log\sigma\parens*{
        \beta \log \frac{\pi'\call{Y_w|X}}{\pi_{\text{ref}}\call{Y_w|X}} - \beta \log \frac{\pi'\call{Y_l|X}}{\pi_{\text{ref}}\call{Y_l|X}}
    }
\bigg]\bigg|\\
\leq{}&  \mathbb{E}_{G}\bigg[\bigg|
    \log\sigma\parens*{
        \beta \log \frac{\pi\call{Y_w|X}}{\pi_{\text{ref}}\call{Y_w|X}} - \beta \log \frac{\pi\call{Y_l|X}}{\pi_{\text{ref}}\call{Y_l|X}}
    }
\nonumber\\&\;\;\;\;\;\;\;\;
    -\log\sigma\parens*{
        \beta \log \frac{\pi'\call{Y_w|X}}{\pi_{\text{ref}}\call{Y_w|X}} 
        - \beta \log \frac{\pi'\call{Y_l|X}}{\pi_{\text{ref}}\call{Y_l|X}}
    }
\bigg|\bigg]\\
\shortintertext{When $\sigma$ is the sigmoid, $\log \sigma$ is concave, so the above is upper bounded:}
\leq{}&  \mathbb{E}_{G}\bigg[\bigg|
    \parens*{
        \beta \log \frac{\pi\call{Y_w|X}}{\pi_{\text{ref}}\call{Y_w|X}} - 
        \beta \log \frac{\pi\call{Y_l|X}}{\pi_{\text{ref}}\call{Y_l|X}}
    }
\nonumber\\&\;\;\;\;\;\;\;\;
    -\parens*{
        \beta \log \frac{\pi'\call{Y_w|X}}{\pi_{\text{ref}}\call{Y_w|X}} -
        \beta \log \frac{\pi'\call{Y_l|X}}{\pi_{\text{ref}}\call{Y_l|X}}
    }
\bigg|\bigg]\\
\leq{}&  \mathbb{E}_{G}\bigg[\bigg|
    \parens*{
        \beta \log \frac{\pi\call{Y_w|X}}{\pi_{\text{ref}}\call{Y_w|X}} -
        \beta \log \frac{\pi'\call{Y_w|X}}{\pi_{\text{ref}}\call{Y_w|X}}
    }
\nonumber\\&\;\;\;\;\;\;\;\;
    -\parens*{
        \beta \log \frac{\pi\call{Y_l|X}}{\pi_{\text{ref}}\call{Y_l|X}} -
        \beta \log \frac{\pi'\call{Y_l|X}}{\pi_{\text{ref}}\call{Y_l|X}}
    }
\bigg|\bigg]\\
\leq{}&  \E[G]{\abs*{
    (r_{\pi}(X, Y_w) - r_{\pi'}(X, Y_w)) - (r_{\pi}(X, Y_l) - r_{\pi'}(X, Y_l))
}}\\
\leq{}&  \E[G]{\abs*{
    (r_{\pi}(X, Y_w) - r_{\pi'}(X, Y_w)) 
}} +
\E[G]{\abs*{
    (r_{\pi}(X, Y_l) - r_{\pi'}(X, Y_l))
}}\\
\leq{}&  2\norm*{r_{\pi} - r_{\pi'}}_{\infty}
\end{align}

Now let $\hat{G}_n$ be the empirical measure of $(X, Y_w, Y_l)$ with $n$ samples. And define

\begin{align}
L_{\hat G_n} \call{\pi} &= R_{\hat G_n}\call{\pi} - R_G\call{\pi}
\end{align}

Then

\begin{align}
\abs*{L_{\hat G_n}\call{\pi} - L_{\hat G_n}\call{\pi'}}
&= \abs*{
    R_{\hat G_n}\call{\pi} -
    R_G\call{\pi} -
    R_{\hat G_n}\call{\pi'} +
    R_G\call{\pi'}
} \\
&\leq \abs*{
    R_{\hat G_n}\call{\pi} -
    R_{\hat G_n}\call{\pi'} + R_G\call{\pi'} - R_G\call{\pi}
} \\
&\leq
    \abs*{R_{\hat G_n}\call{\pi} - R_{\hat G_n}\call{\pi'}} +
    \abs*{R_G\call{\pi'} - R_G\call{\pi}}
\\
&\leq 4 \norm*{r_{\pi} - r_{\pi'}}_{\infty}
\end{align}

So now let $\mathcal K$ be a $\kappa$-cover of $\Pi$ in $d_r(\cdot, \cdot)$. Define the sets $D\call{\pi_k} = \{\pi \in \Pi \; : \; d_r(\pi_k, \pi) \leq \kappa\}$. Then

\begin{align}
P\call{ \sup_{\pi \in \Pi} \abs*{L_{\hat G_n}\call{\pi}} \geq \epsilon } 
&= 
P\call{ \bigcup_{\pi_k \in \mathcal K} \bigg\{ \sup_{\pi \in D\call{\pi_k}} \abs*{L_{\hat G_n}\call{\pi}} \geq \epsilon \bigg\} } \\
&\leq 
\sum_{\pi_k \in \mathcal K} P\call{ \sup_{\pi \in D\call{\pi_k}} \abs*{L_{\hat G_n}\call{\pi}} \geq \epsilon  }
\end{align}

Set $\kappa = \frac{\alpha \epsilon}{4}$ for $0 < \alpha < 1$. Using the above, for any $\pi \in D\call{\pi_k}$ we have

\begin{align}
\abs*{L_{\hat G_n}\call{\pi}  - L_{\hat G_n}\call{\pi_k}} &\leq 4\norm*{r_{\pi} - r_{\pi_k}}_\infty \\
&= 4d_r(\pi_k, \pi) \\
&\leq 4\kappa
\end{align}

By triangle inequality:

\begin{equation}
\abs*{L_{\hat G_n}\call{\pi}}  - \abs*{L_{\hat G_n}\call{\pi_k}} \leq \abs*{L_{\hat G_n}\call{\pi}  - L_{\hat G_n}\call{\pi_k}}
\end{equation}

So

\begin{equation}
\abs*{L_{\hat G_n}\call{\pi}}   \leq 4\kappa + \abs*{L_{\hat G_n}\call{\pi_k}}\\
\abs*{L_{\hat G_n}\call{\pi}}   \leq \alpha\epsilon + \abs*{L_{\hat G_n}\call{\pi_k}}
\end{equation}

Since the probability of the supremum of over a cover is greater than $\epsilon$ implies that the upper bound over the cover is greater than $\epsilon$, we have that the probability of the latter is at least the probability of the former:

\begin{align}
P\call{ \sup_{\pi \in \Pi} \abs*{L_{\hat G_n}\call{\pi}} \geq \epsilon }
&\leq 
\sum_{\pi_k \in \mathcal K} P\call{ \sup_{\pi \in D\call{\pi_k}} \abs*{L_{\hat G_n}\call{\pi}} \geq \epsilon  } \\
&\leq 
\sum_{\pi_k \in \mathcal K} P\call{ \alpha\epsilon + \abs*{L_{\hat G_n}\call{\pi_k}} \geq \epsilon  } \\
&\leq 
\sum_{\pi_k \in \mathcal K} P\call{ \abs*{L_{\hat G_n}\call{\pi_k}} \geq \epsilon(1-\alpha)} \\
\end{align}

Then Hoeffding’s inequality gives

\begin{align}
P\call{ \sup_{\pi \in \Pi} \abs*{L_{\hat G_n}(\pi)} \geq \epsilon }
&\leq 
\sum_{\pi_k \in \mathcal K} P\call{\abs*{L_{\hat G_n}(\pi_k)} \geq \epsilon(1-\alpha)} \\
&\leq 2\abs*{\mathcal K}\exp\call{-\frac{2(1-\alpha)^2n\epsilon^2}{\abs*{2C}^2}}
\end{align}

\end{proof}


\begin{proposition}[Covering number in terms of that of domain and range]\label{prop:covering-number-in-terms-of-domain-range}
    Fix $\cproxy{\phitilde}, \cproxy{\tilde\Theta}, \cproxy{\tautilde^\circ}$ and $L_{\pibar
    }$.
    For $\vec p \in \cproxy{\tautilde^\circ}\call{\set X}$, let 
    \begin{align}
        \bar{\Pi}_{\vec p} &:= \braces*{
            g\call{\cdot_y} = \cproxy{\phitilde}\call{\cproxy{\tilde\Theta} \pibar\call{\vec p}}\bracks{\cdot_y} \;|\; \cproxy{\phitilde}\circ \cproxy{\tilde\Theta}\pibar\circ \cproxy{\tautilde^\circ} 
            \call{\cdot_x} \bracks*{\cdot_y} \in \Pi\call{\cproxy{\phitilde}, \cproxy{\tilde\Theta}, \cproxy{\tautilde^\circ}, L_{\pibar
    }}
        } \subseteq \set{P_Y}
    \end{align}
    and recall the metric on $\set{P_Y}$:
    \begin{align}
        d_{\set{P_Y}} \call{p, q} &:=  \norm*{\beta \log \frac{p\call{\cdot_y}}{q\call{\cdot_y}}}_\infty
    \end{align}

    Then for $\kappa, \delta > 0$, denoting as $\Delta_{\delta}^D$ the $\delta$-cover of $\Delta^D$ under metric $d_{\Delta^D}$:
    {\normalfont
    \begin{align}
        &\texttt{Cov}\call{\Pi, d_r, 3\kappa + 3L_{\phi}\norm{\cproxy{\tilde\Theta}}_p L_{\pibar}\delta } \\
        &\leq \sup_{\vec p' \in \Delta^D_\delta} \texttt{Cov} \call{\bar\Pi_{\vec p'}, d_{\set{P_Y}}, \kappa}^{\texttt{Cov}\call{\Delta^D, d_{\Delta }, \delta}}
    \end{align}
    }
\end{proposition}

\begin{proof}[Proof of Proposition~\ref{prop:covering-number-in-terms-of-domain-range}.]
    We wish to find the covering number of $\Pi$ with the metric $d_r$.
    \paragraph{Take covers of the domain and range of $\pibar$.}
    Take a $\delta$-cover of $ \cproxy{\tautilde^\circ} \call{\set X}$ with metric $d_{\Delta}$, denote it $\Delta^D_{\delta}$. For every $\vec p' \in \Delta^D_\delta$, take a $\kappa$-cover of $\bar\Pi_{\vec p '}$ with metric $d_{\set{P_Y}}$. Denote it by $\bar\Pi_{\vec p ', \kappa}$.

    \paragraph{Construct a set of maps from the domain of $\pibar$ to the range of $\pibar$, via the covers.}
    Let $\bar h_{\Delta^D_\delta}$ be a map from $\Delta^D_\delta$ s.t. for every $\vec p'$, $\bar h_{\Delta^D_\delta}\call{\vec p '} \in \bar\Pi_{\vec p ', \kappa}$. For every such $\bar h_{\Delta^D_\delta}$, extend it to $\bar h$, a function whose domain is $\Delta^D$ as follows: for $\vec p \in \Delta^D$, let: 
    \begin{align}
        \bar h \call{\vec p} &= \begin{cases}
            \bar h_{\Delta_\delta^D} \call{\vec p}, & \text{if }\vec p \in \Delta_\delta^D\\
            \bar h_{\Delta_\delta^D} \call{\vec p'}  \text{ where $\vec p'$ is selected randomly from }A\call{\vec p}, & \text{if }\vec p \not\in \Delta_\delta^D
        \end{cases},\\
        A\call{\vec p} &= \bracesg*{\vec p' \in \Delta_\delta^D}{d_{\Delta^D} \call{\vec p, \vec p '} = \min_{\vec p'' \in \Delta_\delta^D} d_{\Delta^d}\call{\vec p, \vec p''}}.
    \end{align}

    \paragraph{Construct a set of maps $\set X \times \set Y \to \bracks*{0,1}$, denoted by $\set H_{\delta, \kappa}$. This set will be proved to cover $\Pi$.}
    For every $\bar h$, define $h: \set X \times \set Y \to \bracks*{0,1}$:
    \begin{align}
        h\call{x, y} &= \bar h\call{\cproxy{\tautilde^\circ}\call{x}}\bracks*{y} 
    \end{align}
    Let $\set H_{\delta, \kappa}$ denote all such $h$. It can be checked that $\set H_{\delta, \kappa} \subset \Pi$: 
    \begin{enumerate}
        \item $\bar h \call{\cproxy{\tautilde^\circ} \call{x}} \in \bar\Pi_{\vec p'}$ for some $\vec p'$ so is in $\set{P_Y}$; therefore, $\sum_y \bar h \call{\cproxy{\tautilde^\circ} \call{x}}[y] = 1$.
        \item Since $\vec p' \in \cproxy{\tautilde^\circ} \call{\set X}$, take $x'$ so that $\vec p' = \cproxy{\tautilde^\circ}\call{x'}$. Then wlog for every $x$, there is some $\vec p'$ and $x'$ such that,
        \begin{align}
            \bar h \call{\cproxy{\tautilde^\circ} \call{x}}[y] &= \bar h \call{\vec p'}[y]\\
            &= \bar h_{\Delta^D_\delta} \call{\vec p'}[y]\\
            &= \cproxy{\phitilde}\circ\cproxy{\tilde\Theta}\pibar \call{\vec p'}[y]\\
            &= \cproxy{\phitilde}\circ\cproxy{\tilde\Theta}\pibar \call{\cproxy{\tautilde^\circ}\call{x'}}[y],
        \end{align}
        where the third equality holds since $\bar h_{\Delta^D_\delta} \call{\vec p'}[y] \in \bar\Pi_{\vec p'}$. Therefore for every $x'$,  $\abs*{\beta\log\frac{\bar h \call{\cproxy{\tautilde^\circ} \call{x}}[y]}{\pi_{ref}\call{y|x}}} \leq C$.
        \item Now let us show that for some $\pibar_h$, $h = \cproxy{\phitilde}\circ\tilde
        \Theta\pibar_h \circ \cproxy{\tautilde^\circ} \in \mathring{\Pi}$. For every $\vec p \in \cproxy{\tautilde^\circ} \call{\set X}$, $\bar h \call{\vec p} \in \bar\Pi_{\vec p', \kappa}$ for some $\vec p'$, i.e. there is some $\pibar_{\vec p}$ s.t. $\bar h\call{\vec p} = \cproxy{\phitilde}\call{\cproxy{\tilde\Theta} \pibar\call{\vec p'}}$. Construct $\pibar_h\call{\vec p} := \pibar_{\vec p}\call{\vec p'}$. 
    \end{enumerate}

    We need to show that $\set H_{\delta, \kappa}$ covers $\Pi$ with metric $d_r$. To this end we need to show that for every $\pi \in \Pi$, there is a $h_\pi \in \set H_{\delta, \kappa}$, such that $d_r\call{\pi, h_\pi} < \epsilon\call{\delta, \kappa}$ for some small value $\epsilon$ which depends monotonically on $\delta$ and $\kappa$, and $\epsilon\call{0,0}=0$.

    So consider $d_r\call{\pi, h}$ for some $\pi \in \Pi$ and $h \in \set H_{\delta, \kappa}$: 
    \begin{align}
        d_r\call{\pi, h} &= \norm*{
            r_{\pi} - r_h
        }_\infty \\
        &= \sup_{x \in \set X, y \in \set Y} \abs*{
            r_\pi \call{x, y} - r_h\call{x, y}
        }\\
        &= \sup_{x \in \set X, y \in \set Y} \abs*{
            \beta\log\frac{\pi}{\pi_{\text{ref}}} \call{x, y} - \beta\log\frac{h}{\pi_{\text{ref}}} \call{x, y}
        }\\
        &= \sup_{x \in \set X} \braces*{
            \sup_{y \in \set Y}\braces*{
                \abs*{
                    \beta\log\frac{\pi}{\pi_{\text{ref}}} \call{x, y} - \beta\log\frac{h}{\pi_{\text{ref}}} \call{x, y}
                }
            }
        } \\
        &= \sup_{x \in \set X} \braces*{
            \sup_{y \in \set Y}\braces*{
                \abs*{
                    \beta\log{\pi} \call{x, y} - \beta\log{h}\call{x, y}
                }
            }
        } 
    \end{align}
    
    So it is sufficient to show that for every $\pi$ there is some $h_\pi$ such that for every $x$,
    \begin{align}
    &\sup_{y \in \set Y}\braces*{
        \abs*{
            \beta\log{\pi} \call{x, y} - \beta\log{h_\pi}\call{x, y}
        }
    } \\
    &= \norm*{\beta\log{\pi} \call{x, \cdot_y} - \beta\log{h_\pi}\call{x, \cdot_y}}_\infty \\
    &\leq \epsilon\call{\delta, \kappa}
    \end{align}

    \paragraph{Decompose the distance between a general $\pi \in \Pi$ and $h \in \set H_{\delta, \kappa}$ for a fixed $x$. This helps us later choose the $h_\pi$ which makes the bound small enough.}
    Let $\pi \in \Pi$, $h \in \set H_{\delta, \kappa}, x \in \set X$ and $x' \in \set X$ s.t. $\vec p' := \cproxy{\tautilde^\circ} \call{x'} \in A\call{\cproxy{\tautilde^\circ}\call{{x}}}$.
    \begin{align}
        &\norm*{\beta \log \pi \call{x, \cdot_y} - \beta \log h\call{x, \cdot_y}}_\infty \\
        &\leq \norm{\beta \log \pi \call{x, \cdot_y} - \beta \log \pi \call{x', \cdot_y}}_\infty\\
        &+
        \norm{
            \beta \log \pi \call{x', \cdot_y} - \beta \log h\call{x', \cdot_y}
        }_\infty\\
        &+
        \norm{
            \beta \log h\call{x', \cdot_y} - \beta \log h\call{x, \cdot_y}
        }_\infty
    \end{align}

    \paragraph{Consider $\norm*{\beta \log \pi\call{x, \cdot_y} - \beta \log \pi\call{x', \cdot_y}}_\infty$. }
    
        We can show this term is bounded by $L\delta$.

        Let $x' \in \cproxy{\tautilde^{\circ, -1}}\call{\vec p'}$ be s.t. 
        \begin{align}
            &\norm*{\beta \log \pi\call{x, \cdot_y} - \beta \log \pi\call{x', \cdot_y}}_\infty \\
            &\leq 
                \norm*{
                    \beta\log\frac{\pi\call{x, \cdot_y}}{\pi\call{x', \cdot_y}} 
                }_\infty \\
            &\leq 
                d_{\set{P_Y}}\call{\pi\call{x, \cdot_y}, \pi\call{x', \cdot_y}} \\
            &\leq 
                L_{\phi}\norm{\cproxy{\tilde\Theta}}_p L_{\pibar}d_{\Delta}\call{\cproxy{\tautilde^\circ}\call{x}, \cproxy{\tautilde^\circ}\call{x'}} \\
            &\leq 
                L_{\phi}\norm{\cproxy{\tilde\Theta}}_p L_{\pibar}\delta 
        \end{align}

    \paragraph{Consider $\norm*{\beta \log \pi\call{x', \cdot_y} - \beta \log h\call{x', \cdot_y}}_\infty$; we show that we can choose $h_\pi$ to make this be upper bounded by $\kappa$.}
    Since $\pi \in \Pi$, it can be written as $\pi\call{\cdot_x, \cdot_y} = \cproxy{\phitilde}\circ \cproxy{\tilde\Theta}\pibar\circ \cproxy{\tautilde^\circ}\call{\cdot_x}\bracks*{\cdot_y}$ for some $\pibar: \Delta^D \to \Delta^D$. 
    
    Note that $\pi\call{x', \cdot_y}\in \bar \Pi_{\cproxy{\tautilde}\call{x'}}$.
    And since $\cproxy{\tautilde^\circ}\call{x'} \in \Delta_\delta^D$,  there is some $\bar h_{\bar \pi} \in \bar\Pi_{\cproxy{\tautilde^\circ}\call{x'}, \kappa}$ s.t. $d_{\set{P_Y}}\call{\pi\call{x', \cdot_y}, \bar h_{\pibar}\call{\cproxy{\tautilde^\circ}\call{x'}}} \leq \kappa$. But expanding this,
    \begin{align}
        \kappa &\geq d_{\set{P_Y}}\call{\pi\call{
        x', \cdot_y
        }
        , \bar h_{\pibar}\call{\cproxy{\tautilde^\circ}\call{x'}}
        } \\
        &=  \norm*{
            \beta \log \frac{\pi\call{x', \cdot_y}}{\bar h_{\pibar}\call{\cproxy{\tautilde^\circ}\call{x'}}\bracks*{\cdot_y}}
        }_\infty \\
        &= \norm*{
            \beta \log \pi\call{x', \cdot_y} - {\bar h_{\pibar} \circ \cproxy{\tautilde^{\circ}}}\call{x', \cdot_y}
        }_\infty
    \end{align}
    So choose $h_\pi := \bar h_{\pibar} \circ \tautilde^{\circ}$.

    From now on replace $h$ by $h_\pi$.

    \paragraph{Consider $\norm*{\beta \log {h_\pi}\call{x', \cdot_y} - \beta \log {h_\pi}\call{x, \cdot_y}}_\infty$. We show that this is bounded above by $2\kappa + 2L\delta$.}

    Let $\vec p'_1, \vec p'_2 \in A\call{\cproxy{\tautilde^\circ}\call{x}}$, then let $x'_i \in \cproxy{\tautilde^{\circ, -1}}\call{\vec p'_i}$.
    \begin{align}
        \mathrlap{\norm*{
            \beta \log {h_\pi}\call{x'_1, \cdot_y} - \beta \log {h_\pi}\call{x'_2, \cdot_y}        
        }_\infty}\hspace{1cm}&\notag\\
        &\begin{multlined}[b]
            \leq \norm*{
            \beta \log {h_\pi}\call{x'_1, \cdot_y}
            - \beta \log \pi\call{x'_1, \cdot_y}
            }_\infty +{}\\
            \norm*{ \beta \log \pi\call{x'_1, \cdot_y}
            - \beta \log \pi\call{x'_2, \cdot_y}
            }_\infty +{}\\
            \norm*{ \beta \log \pi\call{x'_2, \cdot_y}
            - \beta \log {h_\pi}\call{x'_2, \cdot_y} 
            }_\infty
        \end{multlined}\\
        &\leq \kappa + \norm*{\beta \log \pi\call{x'_1, \cdot_y} - \beta \log \pi\call{x'_2, \cdot_y}}_\infty + \kappa \\
        &= 2\kappa + \beta\norm*{\log\frac{\pi \call{x'_1, \cdot_y}}{\pi\call{x'_2, \cdot_y}} }_\infty\\
        &\leq 2\kappa + L_{\phi}\norm{\cproxy{\tilde\Theta}}_p L_{\pibar}d_{\Delta} \call{\cproxy{\tautilde^\circ} \call{x_1}, \cproxy{\tautilde^\circ} \call{x_2}}  \\
        &\leq 2\kappa + L_{\phi}\norm{\cproxy{\tilde\Theta}}_p L_{\pibar}d_{\Delta } \call{\cproxy{\tautilde^\circ} \call{x_1}, \cproxy{\tautilde^\circ} \call{x}}
        + L_{\phi}\norm{\cproxy{\tilde\Theta}}_p L_{\pibar}d_{\Delta } \call{\cproxy{\tautilde^\circ} \call{x}, \cproxy{\tautilde^\circ} \call{x_2}} \\
        &\leq 2\kappa + 2L_{\phi}\norm{\cproxy{\tilde\Theta}}_p L_{\pibar}\delta 
    \end{align}
    If $\cproxy{\tautilde}\call{x} \not\in \Delta_\delta^D $, $h\call{x} = \bar h\call{\cproxy{\tautilde^\circ}\call{x}} = \bar h_{\Delta_\delta }\call{\vec p''} = h\call{x''}$ with some $\vec p'' \in A\call{\cproxy{\tautilde^\circ}\call{x}}$ and $\cproxy{\tautilde^\circ}\call{x''} = \vec p''$, therefore
    \begin{align}
        \norm*{
            \beta \log {h_\pi}\call{x', \cdot_y} - \beta \log {h_\pi}\call{x, \cdot_y}
        }_\infty
        &=\norm*{
            \beta \log {h_\pi}\call{x', \cdot_y} - r_{h_\pi}\call{x'', \cdot_y}
        }_\infty \\
        &\leq2\kappa + 2L_{\phi}\norm{\cproxy{\tilde\Theta}}_p L_{\pibar}\delta 
    \end{align}

    Therefore, 
    \begin{align}
        \norm*{r_\pi\call{x, \cdot_y} - \beta \log {h_\pi}\call{x, \cdot_y}}_\infty &\leq 3\kappa + 3L_{\phi}\norm{\cproxy{\tilde\Theta}}_p L_{\pibar}\delta 
    \end{align}

    Since the upper bound is constant in $x$, we can conclude that $d_r\call{\pi, h_\pi} \leq 3\kappa + 3L_{\phi}\norm{\cproxy{\tilde\Theta}}_p L_{\pibar}\delta$. So $\set H_{\delta, \kappa}$ covers $\Pi$ in $d_r$ with radius $3\kappa + 3L_{\phi}\norm{\cproxy{\tilde\Theta}}_p L_{\pibar}\delta$.

    Therefore, we have that
    \begin{align}
        \mathrlap{\texttt{Cov}\call{\Pi, d_r, 3\kappa + 3L_{\phi}\norm{\cproxy{\tilde\Theta}}_p L_{\pibar}\delta }}
        &\notag\\
        &\leq \abs*{\set H_{\delta, \kappa}}\\
        &\leq \prod_{\vec p' \in \Delta^D_\delta} \abs*{\bar\Pi_{\vec p', \kappa}} \\
        &\leq \sup_{\vec p' \in \Delta_\delta^D} \texttt{Cov}\call{\bar\Pi_{\vec p'}, d_{\set{P_Y}}, \kappa}^{\texttt{Cov}\call{\Delta^D_\delta, d_{\Delta^D}, \delta}}
    \end{align}
\end{proof}

\subsection{Proof of Theorem~\ref{thm:covering-number-in-terms-of-dimension}}\label{proof:covering-number-in-terms-of-dimension}

\textbf{Theorem~\ref{thm:covering-number-in-terms-of-dimension}}(Bounding sample complexity in terms of dimension)
\emph{
    We remain in the set up of Proposition~\ref{prop:covering-number-in-terms-of-domain-range}. The covering number of $\Pi$ is bounded above by a function of $D$:
    {\normalfont\begin{flalign}
        \mathrlap{\texttt{Cov}\call{\Pi, d_r, 3\kappa + 3L_{\phi}\norm{\cproxy{\tilde\Theta}}_p L_{\pibar}\delta}}
        &\notag\\
         &\leq \parens*{
            \frac{2 L_{\phi}\norm{\cproxy{\tilde\Theta}}_p\sqrt{D}}{\kappa}
        }^{
            D\parens*{
               \frac{2 \sqrt{D}}{\delta} 
            }^D
        }
    \end{flalign}}
    Set $\kappa = \frac{\epsilon}{48}$, we need 
    \begin{align}
    n\call{\epsilon, \omega}
    &= \Omega \call{
            \frac{D}{\epsilon^2}
            \parens*{
               \frac{
                    96L_{\phi}
                    \norm{\cproxy{\tilde\Theta}}_pL_{\pibar} 
               \sqrt{D}
               }{
               \epsilon 
            }
        }^D \log \parens*{
            \frac{96L_{\phi}\norm{\cproxy{\tilde\Theta}}_p \sqrt{D}}{\epsilon}
        }
        - \log \omega
    }
\end{align}
samples to generalise. That is, whenever $n' \geq n(\epsilon, \omega)$, we have 
\begin{align}
    P\bigg(\sup_{\pi \in \Pi} |R_G\call{\pi} - R_{\hat G_{n'}}\call{\pi} | \geq \epsilon \bigg) \leq \omega
\end{align}
}

\begin{proof}[Proof of Theorem~\ref{thm:covering-number-in-terms-of-dimension}.]
    We will bound both $\texttt{Cov}\call{\bar\Pi_{\vec p'}, d_{\set{P_Y}}, \kappa}$ and $\texttt{Cov}\call{\Delta^D, d_{\Delta}, \delta}$ in terms of $D$. 

    First consider $\texttt{Cov}\call{\bar\Pi_{\vec p'}, d_{\set{P_Y}}, \kappa}$. Recall $\bar\Pi_{\vec p'}$:
    \begin{align}
        \bar\Pi_{\vec p'} &= \bracesg*{
            g\call{\cdot_y} = \cproxy{\phitilde}\circ \cproxy{\tilde\Theta} \pibar\call{\vec p' } \bracks*{\cdot_y}}{\cproxy{\phitilde}\circ \cproxy{\tilde\Theta} \pibar\circ \cproxy{\tautilde^\circ}\call{\cdot_{x} } \bracks*{\cdot_y} \in \Pi
        }
    \end{align}

    Now we create a Lipschitz function such that the image is $\bar\Pi_{\vec p'}$: note that $\cproxy{\phitilde}\circ\cproxy{\tilde\Theta}$ is $L_{\phi}\norm{\cproxy{\tilde\Theta}}_p$-Lipschitz, where we recall that $L_{\phi}$ is the Lipschitz constant for $\cproxy{\phitilde}$ and $\norm{\cproxy{\tilde\Theta}}_p$ is the operator-$p$-norm of $\cproxy{\tilde\Theta}$ on $\Delta^D$. For a given $\vec p'$, let $ K(\vec p') = \bracesg*{\pibar\call{\vec p'}}{\cproxy{\phitilde}\circ \cproxy{\tilde\Theta}\pibar\circ \cproxy{\tautilde^\circ}\call{\cdot_x}\bracks*{\cdot_y} \in \Pi} \subseteq \Delta^D$. Then $\cproxy{\phitilde}\circ\cproxy{\tilde\Theta}: K\call{\vec p'} \to \bar\Pi_{\vec p'}$.

    \paragraph{Now use the covering number of $K\call{\vec p'}$ to bound that of $\bar\Pi_{\vec p'}$.}
    
    \begin{align}
        \texttt{Cov}\call{\bar\Pi_{\vec p'}, d_{\set{P_Y}},\kappa} &\leq \texttt{Cov}\call{ K\call{\vec p'}, d_{\Delta}, \frac{\kappa}{L_{\phi}\norm{\cproxy{\tilde\Theta}}_p}} \leq \texttt{Cov}\call{\Delta^D, d_{\Delta}, \frac{\kappa}{L_{\phi}\norm{\cproxy{\tilde\Theta}}_p}}
    \end{align}

    \paragraph{Finally, since all vectors on $\Delta^D$ have bounded $p$-norm, we can bound, for some constant $E\call{p, D}$ depending on the norm:}

    \begin{align}
        \texttt{Cov}\call{\Delta^D, d_{\Delta}, \kappa} &\leq \parens*{\frac{2E\call{p, D}{L_{\phi}\norm{\cproxy{\tilde\Theta}}_p}\sqrt{D}}{\kappa}}^D
    \end{align}
    This gives us:
    \begin{align}
        &\texttt{Cov}\call{\Pi, d_r, 3\kappa + 3L_{\phi}\norm{\cproxy{\tilde\Theta}}_p L_{\pibar}\delta} \\
        &\leq \sup_{\vec p' \in \Delta^D_\delta} \texttt{Cov}\call{\bar\Pi_{\vec p'}, d_{\set{P_Y}}, \kappa}^{\texttt{Cov}\call{\Delta^D, d_{\Delta}, \delta}} \\
        &\leq \texttt{Cov}\call{\Delta^D, d_{\Delta}, \frac{\kappa}{L_{\phi}\norm{\cproxy{\tilde\Theta}}_p }}^{\texttt{Cov}\call{\Delta^D, d_{\Delta}, \delta}} \\ 
        &\leq \parens*{
            \frac{2E\call{p, D}L_{\phi}\norm{\cproxy{\tilde\Theta}}_p\sqrt{D}}{\kappa}
        }^{
            D\parens*{
               \frac{2E\call{p, D}\sqrt{D}}{\delta} 
            }^D
        }
    \end{align}

And let $\kappa = L_{\phi} \norm{\cproxy{\tilde\Theta}}_p \parens*{L_{\pibar}}\delta$.

Then the covering number bound becomes
\begin{align}
        \texttt{Cov}\call{\Pi, d_r, 6L_{\phi} \norm{\cproxy{\tilde\Theta}}_p {L_{\pibar}}\delta}
        &\leq           \texttt{Cov}\call{\Delta^D, d_{\Delta}, {{L_{\pibar} }\delta}}^{\texttt{Cov}\call{\Delta^D, d_{\Delta}, \delta}} \\ 
        &\leq \parens*{
            \frac{2E\call{p, D}\sqrt{D}}{{{L_{\pibar}}\delta}}
        }^{
            D\parens*{
               \frac{2E\call{p, D}\sqrt{D}}{\delta} 
            }^D
        }
    \end{align}

Recall that the generalisation error bound is
    \begin{equation}
    P\bigg(\sup_{\pi \in \Pi} |R_G\call{\pi} - R_{\hat G_n}\call{\pi} | \geq \epsilon \bigg) 
    \leq 
        2 \inf_{\alpha \in \parens*{0,1}} \texttt{Cov} \big( \Pi, d_r, \frac{\alpha \epsilon}{4} \big) e^{-\frac{2\call{1-\alpha}^2n\epsilon^2}{4C^2}}
\end{equation}

For simplicity let $\alpha = \frac{1}{ 2}$. So, set 
    \begin{align}
        \frac{\epsilon} 8 &= 
            6L_{\phi} \norm{\cproxy{\tilde\Theta}}_p L_{\pibar} \delta 
    \end{align}
    So
    \begin{align}
        \delta &=
            \frac{\epsilon}{
            48L_{\phi} \norm{\cproxy{\tilde\Theta}}_p L_{\pibar} 
            }.
    \end{align}
    and
    \begin{align}
        \kappa &=
            \frac{L_\phi \norm{\cproxy{\tilde\Theta}}_p L_{\pibar}\epsilon}{
            48L_{\phi} \norm{\cproxy{\tilde\Theta}}_p L_{\pibar} 
            }\\
            &= \frac{\epsilon}{48}
    \end{align}

    We have
    \begin{align}
        &P\bigg(\sup_{\pi \in \Pi} |R_G\call{\pi} - R_{\hat G_n}\call{\pi} | \geq \epsilon \bigg)\\ 
    &\leq \texttt{Cov}\call{\Pi, d_r, \epsilon/ 8}e^{-\frac{n\epsilon^2}{8C^2}} \\
        &\leq e^{-\frac{n\epsilon^2}{8C^2}}\parens*{
            \frac{96L_{\phi}\norm*{\tilde
            \Theta}_p E\call{p, D}\sqrt{D}}{\epsilon}
        }^{
            D\parens*{
               \frac{
                    96L_{\phi}
                    \norm{\cproxy{\tilde\Theta}}_pL_{\pibar} 
               E\call{p, D}\sqrt{D}
               }{
               \epsilon 
            }
        }^D
    } \label{eq:final-bound-w-proxy}
    \end{align}

For simplices,  $E[p, D] \leq 1$. 

So let's say we want the probility upper bound to be $\omega$, then the number of samples $n$ we need to generalise is
\begin{align}
    \omega &= e^{-\frac{n\epsilon^2}{8C^2}}\parens*{
            \frac{96L_{\phi}\norm*{\tilde
            \Theta}_p \sqrt{D}}{\epsilon}
        }^{
            D\parens*{
               \frac{
                    96L_{\phi}
                    \norm{\cproxy{\tilde\Theta}}_pL_{\pibar} 
               \sqrt{D}
               }{
               \epsilon 
            }
        }^D
    } \\
    n
    &= \Omega \call{
            \frac{D}{\epsilon^2}
            \parens*{
               \frac{
                    96L_{\phi}
                    \norm{\cproxy{\tilde\Theta}}_pL_{\pibar} 
               \sqrt{D}
               }{
               \epsilon 
            }
        }^D \log \parens*{
            \frac{96L_{\phi}\norm*{\tilde
            \Theta}_p \sqrt{D}}{\epsilon}
        }
        - \log \omega
    }
\end{align}

\end{proof}

\subsection{Proof of Theorem~\ref{thm:sample-complexity-without-proxy}}\label{proof:sample-complexity-without-proxy}
\textbf{Theorem~\ref{thm:sample-complexity-without-proxy}} (Bounding sample complexity of learning without proxy)
\emph{
    Let $\mathring\Pi\call{L_{\phi} \norm{\cproxy{\tilde\Theta}}_p L_{\pibar}}$ be the subset of $\mathring\Pi$ where $\pi$ is $L_{\phi} \norm{\cproxy{\tilde\Theta}}_p L_{\pibar}$-Lipschitz.
    Set $\kappa = \frac{\epsilon}{24}$, we need 
    \begin{align}
        \Omega \call{
                \frac{D'}{\epsilon^2}
                \parens*{
                   \frac{
                        48L_{\phi}
                        \norm{\cproxy{\tilde\Theta}}_p{L_{\pibar } }E'\call{p, D'}
                   \sqrt{D'}
                   }{
                   \epsilon 
                }
            }^{D'} \log \parens*{
                \frac{48L_{\phi}\norm{\cproxy{\tilde\Theta}}_p L_{\pibar} E'\call{p, D'} \sqrt{D'}}{\epsilon}
            }
            - \log \omega
        }
    \end{align}
samples to generalise, where $D' \gg D$ , and $E'\call{p, D'} \gg 1$. That is, whenever $n' \geq n(\epsilon, \omega)$, we have 
\begin{align}
    P\bigg(\sup_{\pi \in \mathring\Pi\call{L_{\phi} \norm{\cproxy{\tilde\Theta}}_p L_{\pibar}}} |R_G\call{\pi} - R_{\hat G_{n'}}\call{\pi} | \geq \epsilon \bigg) \leq \delta
\end{align}
}

\begin{proof}[Proof of Theorem~\ref{thm:sample-complexity-without-proxy}]
For learning $\Pi\call{\cproxy{\phitilde}, \cproxy{\tilde\Theta}, \cproxy{\tautilde^\circ}, L_{\pi}}$, the covering number bound is as in Eq~\ref{eq:final-bound-w-proxy}. 

For learning $\mathring{\Pi}$ with the same Lipschitz constant (i.e. $L_{\phi}\norm{\cproxy{\tilde\Theta}}_pL_{\pibar}$) as above but without proxy data, the covering number bound can be read off from \citet{elesedy2022group}. Since $\set X$ is a discrete space, we use the $p$-norm-induced metric in the embedding space of $\set X$; denote the embedding function $f$. Additionally denote the feasible subset in $\set{P_Y}$ by $\set P$. Denote the hypothesis class as $\mathring{\Pi}\call{L_{\phi}\norm{\cproxy{\tilde\Theta}}_pL_{\pibar}}$ to mean the subset with smallest Lipschitz-constant in the argument,
\begin{align}
    &\texttt{Cov}\call{\mathring{\Pi}\call{L_{\phi}\norm{\cproxy{\tilde\Theta}}_pL_{\pibar}}, d_r, 2L_{\phi}\norm{\cproxy{\tilde\Theta}}_p L_{\pibar}\delta + \kappa} \\
    &=
        \texttt{Cov}\call{\set Y, d_{\set{P_Y}}, \kappa}^{\texttt{Cov}\call{
            f\call{\set X}, d_p, \delta
        }} \\
    &\leq \parens*{
            \frac{2E'\call{p, D'}L_{\phi}\norm{\cproxy{\tilde\Theta}}_pL_{\pibar}\sqrt{D'}}{\kappa}
        }^{
            D'\parens*{
               \frac{2E'\call{p, D'}\sqrt{D'}}{\delta} 
            }^{D'}
        }
\end{align}

Setting $\kappa = L_{\phi} \norm{\cproxy{\tilde\Theta}}_p L_{\pibar} \delta$,

\begin{align}
    &\texttt{Cov}\call{\mathring{\Pi}\call{L_{\phi}\norm{\cproxy{\tilde\Theta}}_pL_{\pibar}}, d_r, 2L_{\phi}\norm{\cproxy{\tilde\Theta}}_p L_{\pibar}\delta + \kappa} \\
    &\leq \parens*{
            \frac{2E'\call{p, D'}\sqrt{D'}}{ \delta}
        }^{
            D'\parens*{
               \frac{2E'\call{p, D'}\sqrt{D'}}{\delta} 
            }^{D'}
        }
\end{align}

then setting $\epsilon / 8 = 3L_{\phi} \norm{\cproxy{\tilde\Theta}}_p L_{\pibar} \delta$ (i.e. setting $\kappa = \frac{\epsilon}{24}$)

\begin{align}
    &P\bigg(\sup_{\pi \in \Pi} |R_G\call{\pi} - R_{\hat G_n}\call{\pi} | \geq \epsilon \bigg) \\
    & \leq \texttt{Cov}\call{\mathring{\Pi}\call{L_{\phi}\norm{\cproxy{\tilde\Theta}}_pL_{\pibar}}, d_r, 2L_{\phi}\norm{\cproxy{\tilde\Theta}}_p L_{\pibar}\delta + \kappa}e^{-\frac{n\epsilon^2}{8C^2}} \\
    &\leq e^{-\frac{n\epsilon^2}{8C^2}}\parens*{
            \frac{48L_{\phi} \norm{\cproxy{\tilde\Theta}}_p L_{\pibar}E'\call{p, D'}\sqrt{D'}}{ \epsilon}
        }^{
            D'\parens*{
               \frac{48L_{\phi} \norm{\cproxy{\tilde\Theta}}_p L_{\pibar}E'\call{p, D'}\sqrt{D'}}{\epsilon} 
            }^{D'}
        }
\end{align}

A similar analysis show that we need 
\begin{align}
    n
    &= \Omega \call{
            \frac{D'}{\epsilon^2}
            \parens*{
               \frac{
                    48L_{\phi}
                    \norm{\cproxy{\tilde\Theta}}_p{L_{\pibar } }E'\call{p, D'}
               \sqrt{D'}
               }{
               \epsilon 
            }
        }^{D'} \log \parens*{
            \frac{48L_{\phi}\norm{\cproxy{\tilde\Theta}}_p L_{\pibar} E'\call{p, D'} \sqrt{D'}}{\epsilon}
        }
        - \log \omega
    }
\end{align}

\end{proof}

\end{document}